%% file: acl2023.tex
\pdfoutput=1

\documentclass[11pt]{article}

\usepackage{ACL2023}

\usepackage{times}
\usepackage{latexsym}
\usepackage{enumerate}
\usepackage{enumitem}
\usepackage[T1]{fontenc}

\usepackage[utf8]{inputenc}

\usepackage{microtype}

\usepackage{inconsolata}
\usepackage{amsmath}
\usepackage{dcolumn} 
\usepackage{graphicx}
\usepackage{color}
\usepackage{subcaption}
\usepackage{float}
\usepackage{ifthen}
\usepackage{amssymb}
\newboolean{showcomments}
\setboolean{showcomments}{true}
\ifthenelse{\boolean{showcomments}}
 { \newcommand{\mynote}[2]{
      \fbox{\bfseries\sffamily\scriptsize#1}
        {\small$\blacktriangleright$\textsf{\emph{#2}}$\blacktriangleleft$}}}
        { \newcommand{\mynote}[2]{}}

\title{Testing Low-Resource Language Support in LLMs Using Language Proficiency Exams: the Case of Luxembourgish}

\author{Cedric Lothritz \\
  Luxembourg Institute   \\
  of Science and Technology \\
  5 Av. des Hauts-Fourneaux \\
  L-4362 Esch-sur-Alzette\\
  \texttt{cedric.lothritz@list.lu} \\\And
  Jordi Cabot \\
   Luxembourg Institute   \\
  of Science and Technology \\
   5 Av. des Hauts-Fourneaux \\
  L-4362 Esch-sur-Alzette\\
  \texttt{jordi.cabot@list.lu} \\\And Laura Bernardy\\
  University of Luxembourg\\
  6, rue Coudenhove-Kalergi\\
  L-1359 Luxembourg\\
  \texttt{laura.bernardy@uni.lu}}

\begin{document}
\maketitle
\begin{abstract}
Large Language Models (LLMs) have become an increasingly important tool in research and society at large.
While LLMs are regularly used all over the world by experts and lay-people alike, they are predominantly developed with English-speaking users in mind, performing well in English and other wide-spread languages while less-resourced languages such as Luxembourgish are seen as a lower priority.
This lack of attention is also reflected in the sparsity of available evaluation tools and datasets.
In this study, we investigate the viability of language proficiency exams as such evaluation tools for the Luxembourgish language. 
We find that large models such as Claude and DeepSeek-R1 typically achieve high scores, while smaller models show weak performances. 
We also find that the performances in such language exams can be used to predict performances in other NLP tasks in Luxembourgish.

\end{abstract}

\input{1_introduction}
\input{7_related_works}

\input{4_methodology}

\input{5_results}

\input{6_discussion}
\input{8_conclusion}
\input{09_acknowledgements}
\input{10_limitations}

\bibliography{custom}
\bibliographystyle{acl_natbib}

\appendix
\input{10_appendix}

\end{document}

%% file: 1_introduction.tex
\section{Introduction}
Large Language Models (LLMs) have become increasingly ubiquitous in recent research, with new models being frequently released and updated.
They have become similarly omnipresent in the everyday lives of private and professional users, with models being deployed in wide-ranging domains such as finance, medicine, and coding. 

While LLMs are primarily catered towards speakers of widespread languages such as English, Spanish, and Chinese, low-resource languages are generally regarded as a lower priority.
Indeed, many major LLMs such as the Llama family~(\citealp{touvronllama}, \citealp{touvron2023llama}, \citealp{grattafiori2024llama}) are trained nearly exclusively on English data, and even LLMs with a focus on non-English languages such as BLOOM~\cite{le2023bloom} barely include low-resource languages.
As a consequence, numerous mistakes can occur due to cross-lingual transfer between related languages. 
These mistakes include mistranslated words and partial code-switching, incorrect word order, misgendered nouns, and violation of language-exclusive rules such as the n-rule (Eifeler Regel) in the Luxembourgish language~\cite{gilles2006phonologie}.

As such, it is important to determine the capabilities of models to understand and generate texts in such languages, and to recognise when they can fail. 
The benefits of such research are two-fold as it allows (1) to find general weaknesses and pitfalls that can be addressed and mitigated in future LLMs; (2) to make recommendations on which LLM to use for a given scenario in a given low-resource language.
The latter is especially important in use cases where an LLM would have to handle personally identifiable data such as personal names or phone numbers that is not allowed to leave a company or organisation due to internal policies and data privacy laws such as the Luxembourgish Data Protection Law~\cite{luxdata2018} or the General Data Protection Regulation~\cite{regulation2016regulation}.
In such cases, data must be processed using locally deployed models that do not access the internet.

In this study, we aim to investigate the capabilities of LLMs for Luxembourgish, a low-resource language spoken by nearly 600,000 people world-wide\footnote{\url{https://cursus.edu/en/23040/luxembourgish-at-its-best}}, but predominantly in Luxembourg.
We want to determine those capabilities from two angles, (a) how well LLMs perform in linguistic proficiency exams used in language institutions; (b) whether their performance in such exams can predict their performance in Natural Language Generation (NLG) tasks.

Specifically, we investigate 53 LLMs from a wide array of LLM families that we compare systematically. 
In addition, we ascertain in which cases LLMs fail and discover trends and common weaknesses using language exams. 
Finally, we aim to determine whether or not such exams are a useful tool to predict how well an LLM performs in two tasks related to text summarisation. 
While this study focuses on the Luxembourgish language, we believe that our methodology can be generalised to other low-resource languages as well.

In summary, our contributions are as follows:
\begin{enumerate}[label=(\alph*)]
    \item A systematic comparison of 53 popular LLMs with regard to their linguistic proficiency in Luxembourgish.
    \item A quantitative and qualitative error analysis.
    \item An analysis on the usefulness of language exams as a tool for performance prediction on NLG tasks for Luxembourgish.
    \item An accessible and interactive public leaderboard for performance in Luxembourgish that we will extend as new LLMs are released.\footnote{The leaderboard can be found as part of our AI Sandbox at \url{https://ai-sandbox.list.lu/}}
\end{enumerate}

%% file: 7_related_works.tex
\section{Related Work}
\subsection{Testing LLMs on Knowledge Exams}
Many recent studies have used exams to test the knowledge of LLMs in a variety of domains. 
These domains predominantly cover the medical domain (\citealp{gotta2024large}, \citealp{stribling2024model}, \citealp{csahin2025current}, \citealp{abbas2024comparing}, \citealp{zong2024large}), but also computer science and programming (\citealp{dinh2024sciex}, \citealp{ellis2024chatgpt}, \citealp{li2025investigating},\citealp{varastehnezhad2024llm}), and the legal domain (\citealp{katz2024gpt}, \citealp{martinez2024re}). 
The general consensus is that LLMs perform well in these exams, often surpassing human candidates.
However, most of these studies cover exams written in English, with only a few covering exams from Non-English speaking countries.
\citet{koto2024cracking} released IndoCareer, an exam benchmark covering various topics in Indonesian. They test it on a large number of LLMs with mixed results.
\citet{koto2023large} also conducted a similar study using Indonesian school exams, where they found that most LLMs managed to only pass primary school exams.
\citet{jassem2025llmzsz} presented LLMzSzŁ, a benchmark of Polish national exams covering various subjects.
They test numerous LLMs in both English and Polish, showing that most LLMs perform better than randomly guessing.
\citet{locatelli2024examining} tested LLMs on a standardised entry exam for Brazilian universities covering different high school subjects. They found that all LLMs performed very well on the exams, however, the number of tested LLMs was very low compared to our study.
\citet{dao2023evaluation} compared two LLMs using Vietnamese high school physics exams, showing that they perform worse than students.
\subsection{Testing Linguistic capabilities of LLMs with Language exams}
While there are numerous studies on the performance of LLMs on general NLP tasks, they are far less commonly focused specifically on language exams. 
\citet{dargis2024evaluating} systematically test numerous LLMs on standardised high school language tests in Latvian and manually evaluate their outputs with a human expert. 
\citet{mayor2024evaluating} tested several LLMs on TELEIA, a benchmark of exams for Spanish as a second language, as well as a Spanish test for foreign university students. 
Similarly, \citet{mercorio2024disce} test numerous LLMs on the novel Italian language exam benchmark INVALSI, finding that large models consistently outperform smaller ones, and that the performance degrades on harder tests.

However, compared to related studies, we do not only assess LLM performance on language exams, but we also explore their usefulness to predict LLM performance in general NLP tasks.
Furthermore, we use tests that follow the Common European Framework of Reference for Languages (CEFR)~\cite{council2001common} to evaluate LLMs.

%% file: 4_methodology.tex
\section{Methodology}
In this section, we give details on the methodology of this study and on our experiments, including the datasets we used, the LLMs we targeted, the metrics to compare the LLMs, the prompt design, and hyperparameters of the studied models. We also present and elaborate on the research questions we address in this paper.

We evaluate LLMs in two general disciplines: (1) language fluency and (2) language generation.
The language fluency test consists of solving language exams that comprise sets of multiple-choice questions that relate to the general understanding of the language itself. 
In order to evaluate NLG, we use two tasks from the LuxGen benchmark~\cite{plum2025text}, specifically the Headline Generation and Short Description tasks.
These tasks both consist of processing texts and generating appropriate summaries, either in form of a short description or a news headline.

\begin{figure}
    \centering
    \includegraphics[width=0.4\textwidth]{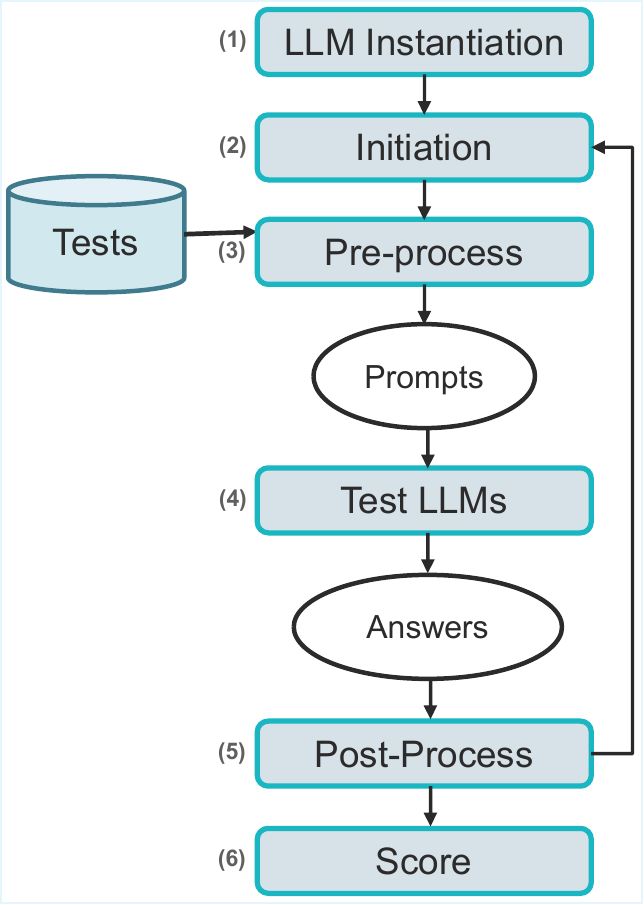}
    \caption{General testing pipeline.}
    \label{fig:pipeline}
\end{figure}

Testing is done in a zero-shot fashion for all tasks. 
Figure~\ref{fig:pipeline} shows the general testing pipeline.
After instantiating an LLM (1), for every test sample, we initiate the model with a prompt that is identical across every LLM (2). 
This initiation prompt consists of explaining the LLM's role and the general task it needs to solve, as well as specifying the format of the inputs and outputs.
We then pre-process the test samples to make them uniform and compatible with our prompts (3), and give them to the LLM one at a time and in random order (4). 
For the language exams, we apply a post-processing step before comparing it to the ground truth (5). 
Post-processing generally consists of automatically correcting minor typos if the output is different from each of the provided answers (however, we judge outputs as being invalid if they deviate significantly or are ambiguous and could be attributed to multiple answers).
Finally, the LLM's performance will be measured using task-specific metrics (6) (see Section~\ref{sec:metrics}).

\subsection{Research Questions}
In this study, we aim to answer the following three research questions:
\begin{enumerate}[leftmargin=0pt,itemsep=0pt]
\item[] \textbf{RQ1: How well do LLMs perform in Luxembourgish language tests?} To answer this question, we let LLMs solve language exams. We compare their performances in terms of number of correctly answered questions and determine a "winner".
\item[] \textbf{RQ2: In which cases do LLMs fail to understand Luxembourgish?} For this question, we categorise the exam questions to determine which categories are more difficult for LLMs to solve.
\item[] \textbf{RQ3: Is there a correlation between performing well in a language exam and performing well in an NLG task?} For the final question, we determine whether using language exams is an appropriate method to decide an LLM's language capabilities and if the "winners" of the exams manage to perform well in generation tasks.
\end{enumerate}

\subsection{Datasets}

\subsubsection{Language Tests}

\begin{table}[t!]
    
    \centering
    \resizebox{\linewidth}{!}{
    \begin{tabular}{|r|l|l|l|l|l|l|l|}
    \hline
        Category & A1 & A2 & B1 & B2 & C1 & C2 & Total \\\hline
         Vocabulary &26 &26&25&28&24&26&155 \\
         Grammar &26 &26&26&34&29&23&164 \\
         Reading C. &26 &26&25&26&26&26&155 \\
         Conversation C. &26 &26&26&26&25&26&155 \\\hline
         Total &104 &104&102&114&104&101&629\\\hline
    \end{tabular}}
    \caption{Statistics for language tests.}
    \label{tab:stat_lang_tests}
\end{table}

\label{sec:language_tests}
In order to evaluate the fluency of each model, we use official language exams for Luxembourgish provided to us by the Institut National des Langues Luxembourg (INLL), a language institute in Luxembourg\footnote{\url{https://www.inll.lu/en/}}. 
The difficulty level of each test follows CEFR guidelines, and thusly ranges from A1 (beginner) to C2 (proficient). 
The tests consist of fill-in-the-blank type questions. The large majority of them are multiple-choice questions (MCQs) while nearly 17\% of them them were \textit{open} without a pre-determined selection of possible answers. 
The number of open questions differs depending on the CEFR level and are either related to grammar or vocabulary. Appendix~\ref{sec:app_pre_a} shows the exact number and distribution of open questions in the exams. The A1 and A2 levels do not contain any open questions, then, from B1 level, the higher the level the higher the number of open questions. 

However, for our experiments, we unify the exams by transforming them into the same type of questions.
Two Luxembourgish-speaking authors converted the open questions into MCQs as this allows to use a single standardised prompt across all tests, and simplifies the error analysis.
The designated authors made sure that the difficulty of the converted questions is adequate for the level of the corresponding exams by writing distractors that matched the difficulty of the given exam, taking other MCQ questions from the same level as inspiration. 
Distractors for grammar-related questions include lexically similar misspellings of the correct answer, incorrectly conjugated verbs, and incorrect usage of prepositions.
Vocabulary-related distractors generally include lexically similar words to the correct answer and near-synonyms to the correct answer that are wrong in the context of the given text. 
Questions fall into one of four broad categories: vocabulary, grammar, reading comprehension (RC), and listening comprehension. 
Note that the data for the latter category consists of transcripts rather than audio files. 
As such, we will refer to that category as conversation comprehension (CC) in this paper. 
Table~\ref{tab:stat_lang_tests} shows further details for the tests.
Further note that we do not have permission to give access to the actual dataset, but we can establish contact between interested researchers and the INLL.

\subsubsection{News Headline Generation}
The Headline Generation task, being part of \citeposs{plum2025text} LuxGen benchmark for assessing Luxembourgish language models, consists of generating an appropriate title for a given news article. 
The articles were sourced from the Luxembourgish news agency RTL. The original test set comprises 13\,852 samples, but for our purposes, we use a subset of 300 randomly selected samples due to the vast number of LLMs we are investigating.  

\subsubsection{Short Description Generation}
The second LuxGen task we consider involves generating short descriptions of Luxembourgish Wikipedia articles. The descriptions featured on Wikidata serve as a ground truth for this task. The original test set consists of 2094 articles, but similarly to the Headline Generation task, we limit ourselves to a random selection of 300 samples.

\begin{table*}[ht!]
    \centering
     \resizebox{\textwidth}{!}{
    \begin{tabular}{|l|l|r|r|r|r|r|r||l|l|r|r|r|r|r|r|}
    \hline
    Model & Size & A1 & A2& B1 & B2& C1 & C2&Model & Size& A1& A2& B1& B2& C1& C2\\\hline
    \multicolumn{8}{|c||}{\textbf{Baseline (average chance rate)}}&\multicolumn{8}{c|}{\textbf{Baseline (average chance rate)}}\\\hline
    Random Baseline&/&37.6&37.5&33.3&33.0&31.8&33.5&Random Baseline&/&37.6&37.5&33.3&33.0&31.8&33.5\\\hline		
    \multicolumn{8}{|c||}{\textbf{ Small Models} ($\leq 15$B)}&\multicolumn{8}{c|}{\textbf{ Medium-sized Models} ($>15$B$  \&  \leq 200$B)}\\\hline
Qwen 2	&	0.5	&	26.9	&	22.1	&	23.3	&	22.8	&	22.1	&	15.8	&Mistral-Small	&	22	&	51.9	&	46.2	&	31.1	&	39.5	&	38.5	&	25.7	\\
Llama 3.2	&	1	&	26.0	&	19.2	&	28.2	&	21.1	&	19.2	&	12.9	&Gemma 2	&	27	&	66.3	&	63.5	&	54.4	&	45.6	&	50.0	&	41.6	\\
Gemma 3	&	1	&	45.2	&	36.5	&	38.8	&	40.4	&	25.0	&	33.7	&Gemma 3	&	27	&	\textbf{71.2}	&	\textbf{72.1}	&	\textbf{62.1}	&	\textbf{60.5	}&	\textbf{60.6	}&	\textbf{44.6}	\\
Qwen 2	&	1.5	&	32.7	&	27.9	&	25.2	&	31.6	&	21.2	&	23.8	&QWQ	&	32	&	60.6	&	49.0	&	51.5	&	47.4	&	38.5	&	\textbf{44.6}	\\
StableLM 2	&	1.6	&	22.1	&	14.4	&	21.4	&	16.7	&	15.4	&	16.8	&Aya-23	&	35	&	43.3	&	37.5	&	39.8	&	36.0	&	33.7	&	34.7	\\
EuroLLM	&	1.7	&	17.3	&	21.2	&	12.6	&	23.7	&	26.0	&	17.8	&Command-R	&	35	&	51.0	&	38.5	&	33.0	&	42.1	&	35.6	&	37.6	\\
Gemma 2	&	2	&	39.4	&	32.7	&	31.1	&	32.5	&	27.9	&	28.7	&Alfred	&	40	&	42.3	&	39.4	&	29.1	&	29.8	&	36.5	&	29.7	\\
Llama 3.2	&	3	&	37.5	&	35.6	&	26.2	&	35.1	&	30.8	&	24.8	&Mixtral	&	8x7	&	50.0	&	41.3	&	34.0	&	39.5	&	31.7	&	29.7	\\
Phi 3	&	3.8	&	45.2	&	36.5	&	27.2	&	26.3	&	27.9	&	17.8	&Llama 2	&	70	&	38.5	&	33.7	&	28.2	&	39.5	&	26.0	&	39.6	\\
Phi 3.5	&	3.8	&	39.4	&	27.9	&	11.7	&	10.5	&	7.7	&	10.9	&Llama 3	&	70	&	52.9	&	49.0	&	38.8	&	35.1	&	38.5	&	43.6	\\
Gemma 3	&	4	&	54.8	&	43.3	&	39.8	&	46.5	&	36.5	&	31.7	&Llama 3.1	&	70	&	55.8	&	56.7	&	49.5	&	52.6	&	47.1	&	41.6	\\
Qwen 2	&	7	&	48.1	&	36.5	&	32.0	&	36.0	&	39.4	&	31.7	&DeepSeek-R1	&	70	&	64.4	&	51.0	&	56.3	&	54.4	&	49.0	&	41.6	\\
Llama 2	&	7	&	36.5	&	38.5	&	24.3	&	39.5	&	25.0	&	26.7	&Qwen 2	&	72	&	60.6	&	46.2	&	40.8	&	51.8	&	41.3	&	40.6	\\
WizardLM 2	&	7	&	44.2	&	38.5	&	21.4	&	31.6	&	28.8	&	29.7	&Command-R+	&	104	&	52.9	&	38.5	&	39.8	&	44.7	&	31.7	&	32.7	\\
Mistral	&	7	&	35.6	&	35.6	&	34.0	&	30.7	&	34.6	&	26.7	&Mistral-Large	&	123	&	66.3	&	50.0	&	42.7	&	52.6	&	43.3	&	\textbf{44.6}	\\
Aya-23	&	8	&	40.4	&	30.8	&	30.1	&	31.6	&	31.7	&	31.7	&Mixtral	&	8x22	&	62.5	&	49.0	&	41.7	&	42.1	&	39.4	&	39.6	\\
Llama 3.1	 & 	8	&	56.7	&	33.7	&	33.0	&	33.3	&	30.8	&	33.7	&WizardLM 2	&	8x22	&	58.7	&	40.4	&	40.8	&	43.0	&	35.6	&	34.7	\\\cline{9-16}
Llama 3	&	8	&	45.2	&	35.6	&	35.0	&	36.0	&	28.8	&	31.7	&\multicolumn{8}{c|}{\textbf{ Large Models} ($> 200$B)}\\\cline{9-16}															
DeepSeek-R1	&	8	&	53.8	&	39.4	&	38.8	&	39.5	&	34.6	&	28.7	&Llama 3.1	&	405	&	78.8	&	39.4	&	48.5	&	59.6	&	55.8	&	46.5	\\

GLM 4	&	9	&	54.8	&	41.3	&	37.9	&	45.6	&	44.2	&	35.6	&DeepSeek-R1	&	671	&	\textbf{92.3	}&	83.7	&	73.8	&	76.3	&	73.1	&	\textbf{81.2}	\\
EuroLLM	&	9	&	44.2	&	45.2	&	33.0	&	34.2	&	31.7	&	25.7	&ChatGPT 3.5	&	unk.	&	\textbf{92.3	}&	85.6	&	\textbf{80.6}	&	83.3	&	76.0	&	66.3	\\
Gemma 2	&	9	&	58.7	&	47.1	&	38.8	&	49.1	&	40.4	&	34.7	&Claude 3.5 Sonnet	&	unk.	&	\textbf{92.3}	&	\textbf{92.3}	&	\textbf{80.6}	&	\textbf{86.8}	&	\textbf{77.9	}&	69.3	\\
		Mistral-Nemo	&	12	&	52.9	&	36.5	&	28.2	&	33.3	&	29.8	&	20.8	&ChatGPT 4o	&	unk.	&	\textbf{92.3}	&	85.6	&	78.6	&	84.2	&	74.0	&	60.4	\\
StableLM 2	&	12	&	27.9	&	35.6	&	26.2	&	28.9	&	27.9	&	23.8	&ChatGPT 4o mini	&	unk.	&	90.4	&	90.4	&	59.2	&	59.6	&	54.8	&	43.6	\\
Gemma 3	&	12	&	\textbf{62.5	}&	\textbf{58.7}	&	\textbf{53.4	}&	\textbf{50.9}	&	\textbf{47.1	}&	\textbf{40.6	}&Gemini 2.0 Flash	&	unk.	&	89.4	&	84.6	&	74.8	&	80.7	&	76.0	&	54.5	\\
		Llama 2	&	13	&	27.9	&	14.4	&	23.3	&	26.3	&	23.1	&	26.7	&LeChat	&	unk.	&	42.3	&	53.8	&	38.8	&	51.8	&	41.3	&	32.7	\\\cline{9-16}	
Phi 3	&	14	&	45.2	&	38.5	&	38.8	&	39.5	&	27.9	&	21.8	&\multicolumn{8}{c|}{}\\															
Phi 4	&	14	&	53.8	&	51.9	&	42.7	&	44.7	&	34.6	&	33.7	&\multicolumn{8}{c|}{}\\\hline

    \end{tabular}}
    \caption{Total results of language exams. \textbf{Bold} numbers indicate highest performance for each size category.}
    \label{tab:results_exams_total}
\end{table*}
\subsection{Models}
We perform our experiments on a large number of LLMs from a wide array of language model families and model sizes. 
As models are not explicitly reported to understand Luxembourgish, the main criteria for including a given LLM in our selection is their reported ability to (a) understand a closely related Germanic language such as German or Dutch, or (b) understand a very large number of languages. 
Specifically, we include the following models: Alfred~\cite{alfred-40b-1023}, Aya-23~\cite{aryabumi2024aya}, ChatGPT (\citealp{OpenAI2023ChatGPT3.5}, \citealp{OpenAI2024ChatGPT4o}, \citealp{OpenAI2024ChatGPT4oMini}), Claude 3.5 Sonnet~\cite{claude2024}, Command-R~\cite{cohere_for_ai_2024}, DeepSeek-R1~\cite{deepseekai2025deepseekr1incentivizingreasoningcapability}, EuroLLM~\cite{martins2024eurollm}, Gemini 2.0 Flash~\cite{google_gemini_flash}, LeChat~\cite{LeChat}, Gemma 2~\cite{gemma_2024}, GLM 4~\cite{glm2024chatglm}, Gemma 3~\cite{team2025gemma}, Llama 2~\cite{touvron2023llama}, Llama 3, Llama 3.1, Llama 3.2~\cite{grattafiori2024llama}, Mistral~\cite{Mistral7B}, Mixtral~\cite{Mixtral8x7B}, Phi 3, Phi 3.5~\cite{abdin2024phi}, Phi 4~\cite{abdin2024phi4}, Qwen 2~\cite{qwen2}, QWQ~(\citealp{qwen2.5}, \citealp{qwq32b}),  StableLM 2~\cite{bellagente2024stable}, and WizardLM 2~\cite{xu2024wizardlm}. 
Appendix~\ref{sec:app_a} shows the full list of LLMs together with model sizes and reasons for inclusion.

Model sizes range from 0.5 billion parameters (Qwen2:0.5B) to 671 billion (DeepSeek-R1:671B). 
We divide them into 3 size categories: \emph{small} ($\leq 15$B parameters), \emph{medium-sized} ($>15$B \& $\leq 200$B parameters), and \emph{large} ($> 200$B parameters). 
Note that although some model sizes are unknown to us as they are not communicated by the companies releasing them, they are estimated to be at least hundreds of billions of parameters. 
As such, we categorise the concerned LLMs as \emph{large}.
\subsection{Metrics}
\label{sec:metrics}
We measure the performance of LLMs using different metrics depending on the task.
For the language exams, we use simple accuracy, i.e., the percentage of questions that an LLM answers correctly.

For language generation, we evaluate the outputs of LLMs in terms of grammar and spelling as well as similarity to the ground truth.
We measure the performance on both of these tasks in an LLM-as-a-judge (LaaJ) fashion using an LLM that is not part of the ones we are investigating as the judging LLM.
We choose the judging LLM based on its ability to solve the task at hand, which we manually verify.
The metrics should meet the following requirements:
\begin{itemize}[leftmargin=0pt,itemsep=-5pt]

    \item[] \textit{Grammar and Spelling metric}
\begin{enumerate}[leftmargin=15pt,itemsep=0pt]
    \item range between $0$ and $1$ where $0$ is incomprehensible and $1$ is perfect with neither spelling nor grammar mistakes.
    \item punish non-Luxembourgish outputs, i.e., return $0$ if the output is written in a different language.
\end{enumerate}
\vspace{10pt}
    \item[] \textit{Adequacy metric}
\begin{enumerate}
    \item range between $0$ and $1$ where $1$ is identical to the ground truth and decreases the less similar the output is to the ground truth and the more it deviates from the text on a semantic level.
    \item punish descriptions that are deemed too long.
\end{enumerate}
\end{itemize}


Considering that such LLM-generated metrics are essentially black boxes and non-deterministic due to the inherent random nature of LLMs, we repeat the generation a total of three times in order to reduce variation.
In addition to the LaaJ metrics, we use two traditional metrics: METEOR~\cite{banerjee2005meteor} and BERTScore~\cite{zhang2019bertscore}, using the Luxembourgish LuxemBERT~\cite{lothritz2022luxembert} as the embedding model for BERTScore.

\subsection{Prompt Design}

We experimented with prompts of different lengths and degrees of complexity to find a prompt that yields the best result for solving the language exams. See Appendix~\ref{sec:app_a1} for the full prompt. 
Regarding the prompts for the Headline Generation and Short Description summary tasks, we reuse the prompts used by \newcite{plum2025text} for every model we investigate.
The prompts used for evaluating the LuxGen tasks that use the metric requirements described in Section~\ref{sec:metrics} can be found in Appendix~\ref{sec:app_b}.

\subsection{Computational Setup and Hyperparameters}
LLMs are either accessed externally via API key or stored locally on our HPC infrastructure where we have access to three NVIDIA Tesla V100 32GB GPUs. We use Ollama~\footnote{\url{https://github.com/ollama/ollama}} to prompt the LLMs.

For all our experiments, models and metrics, we use the default hyperparameters. The exceptions are the models solving the language exams, for which we set the temperature to 0 to reduce randomness of the answers.

%% file: 5_results.tex
\section{Results}

\subsection{RQ1: How well do LLMs perform in Luxembourgish language tests?}
\label{sec:rq1}
Table~\ref{tab:results_exams_total} shows the overall results of the LLMs we tested as well as a baseline determined as the average performance for a model that randomly guesses the answers of the exams. See Appendix~\ref{sec:app_c} for a breakdown of results across the four categories. 
For a more convenient way of exploring the results, we also release an interactive web app with the results.

We observe a clear trend where the \emph{large} models typically perform well above average. 
Claude 3.5 Sonnet achieves the highest scores across every test except for the C2 test for which DeepSeek-R1:671B performs best.
Among the \emph{small} models, many models show promise to be adequate for further fine-tuning on Luxembourgish data despite their small size as they perform better than the random baseline and either approach or slightly surpass the 50\% mark on the easier tests. 
We deem Gemma3:12B the best \emph{small} model as it performs best in all exams out of the models in the \emph{small} category.

\emph{Medium-sized} models generally perform consistently better than \emph{small} models.
Gemma3:27B is the overall winner among the \emph{medium-sized} models, performing significantly better than all of its counterparts except for the C2 exams where it ties with the QWQ and Mistral-Large models.

Furthermore, as expected, performances drop for higher CEFR levels as the tests become increasingly more difficult, with one notable exception. 
Most models achieve a higher performance on the B2 test than on the B1 test.
Specifically, they perform surprisingly well on the Reading Comprehension questions, warranting a closer examination of the composition of the exams.

\subsection{RQ2: In which cases do LLMs fail to understand Luxembourgish?}

\begin{table}[ht!]
    \centering
    
\resizebox{.45\textwidth}{!}{    \begin{tabular}{|l|l|r|r|r|r|}
    \hline
model	&	Size	&	Vocab.	&	Grammar	&	RC	&	CC	\\\hline
\multicolumn{6}{|c|}{\textbf{ Small Models ($\leq 15B $)}}\\\hline
Qwen 2	&	0.5	&	25.7	&	19.7	&	21.2	&	22.6	\\
Llama 3.2	&	1	&	21.3	&	23.7	&	20.5	&	18.8	\\
Gemma 3	&	1	&	27.4	&	37.8	&	41.7	&	39.9	\\
Qwen 2	&	1.5	&	29.6	&	28.3	&	25.0	&	25.2	\\
StableLM 2	&	1.6	&	23.2	&	28.5	&	5.8	&	13.5	\\
EuroLLM	&	1.7	&	24.0	&	20.0	&	16.7	&	18.1	\\
Gemma 2	&	2	&	34.4	&	29.7	&	29.5	&	34.8	\\
Llama 3.2	&	3	&	31.7	&	28.3	&	32.1	&	35.5	\\
Phi 3	&	3.8	&	27.8	&	25.3	&	29.5	&	38.0	\\
Phi 3.5	&	3.8	&	17.3	&	20.0	&	15.4	&	19.3	\\
Gemma 3	&	4	&	46.2	&	32.9	&	43.6	&	47.1	\\
Qwen 2	&	7	&	35.8	&	31.2	&	35.9	&	47.1	\\
Llama 2	&	7	&	30.7	&	26.7	&	35.3	&	34.8	\\
WizardLM 2	&	7	&	35.9	&	32.3	&	28.8	&	32.8	\\
Mistral	&	7	&	31.2	&	28.3	&	34.6	&	38.1	\\
Aya-23	&	8	&	31.7	&	29.5	&	32.1	&	37.4	\\
Llama 3.1	&	8	&	30.1	&	33.5	&	39.7	&	44.4	\\
Llama 3	&	8	&	34.5	&	33.7	&	35.3	&	38.6	\\
DeepSeek-R1	&	8	&	40.6	&	34.5	&	40.4	&	41.3	\\
GLM 4	&	9	&	50.3	&	27.3	&	43.6	&	52.8	\\
EuroLLM	&	9	&	40.0	&	32.0	&	35.9	&	35.5	\\
Gemma 2	&	9	&	57.3	&	34.1	&	44.9	&	43.8	\\
Mistral-Nemo	&	12	&	36.2	&	27.7	&	34.6	&	36.7	\\
StableLM 2	&	12	&	35.1	&	17.1	&	30.8	&	31.7	\\
Gemma 3	&	12	&	\textbf{63.1}	&	32.4	&	\textbf{54.5}	&	\textbf{59.9}	\\
Llama 2	&	13	&	25.6	&	26.6	&	21.2	&	21.3	\\
Phi 3	&	14	&	34.0	&	35.8	&	32.7	&	38.6	\\
Phi 4	&	14	&	46.2	&	\textbf{36.7}	&	45.5	&	45.8	\\
\hline											
\multicolumn{6}{|c|}{ \textbf{Medium-sized Models ($>15B \& \leq 200B $)}}\\\hline											
Mistral-Small	&	22	&	41.3	&	29.5	&	34.6	&	50.3	\\
Gemma 2	&	27	&	60.1	&	36.7	&	57.1	&	62.0	\\
Gemma 3	&	27	&	\textbf{71.1}	&	\textbf{51.7}	&	\textbf{60.9}	&	\textbf{64.5}	\\
QWQ	&	32	&	59.5	&	38.1	&	41.7	&	56.0	\\
Aya-23	&	35	&	39.6	&	30.7	&	40.4	&	40.0	\\
Command-R	&	35	&	39.4	&	34.4	&	44.2	&	41.3	\\
Alfred	&	40	&	34.3	&	32.3	&	34.6	&	36.8	\\
Mixtral	&	8x7	&	47.1	&	27.3	&	36.5	&	40.7	\\
Llama 2	&	70	&	28.8	&	28.3	&	36.5	&	43.1	\\
Llama 3	&	70	&	53.0	&	35.6	&	40.4	&	43.8	\\
Llama 3.1	&	70	&	60.4	&	46.0	&	45.5	&	50.9	\\
DeepSeek-R1	&	70	&	64.5	&	42.2	&	44.9	&	60.6	\\
Qwen 2	&	72	&	51.6	&	36.5	&	48.7	&	51.5	\\
Command-R+	&	104	&	46.5	&	31.1	&	40.4	&	42.5	\\
Mistral-Large	&	123	&	64.9	&	41.1	&	42.3	&	52.2	\\
Mixtral	&	8x22	&	49.4	&	40.5	&	42.3	&	50.9	\\
WizardLM 2	&	8x22	&	40.4	&	37.7	&	41.7	&	49.0	\\\hline
\multicolumn{6}{|c|}{ \textbf{Large Models} ($> 200B $)}\\\hline											
Llama 3.1	&	405	&	63.4	&	43.0	&	54.5	&	59.3	\\
DeepSeek-R1	&	671	&	\textbf{92.2	}&	75.2	&	74.4	&	79.9	\\
ChatGPT 3.5	&	unk.	&	91.6	&	72.1	&	73.7	&	85.8	\\
Claude 3.5 Sonnet	&	unk.	&	87.8	&	\textbf{82.4}	&	\textbf{76.9}	&	\textbf{86.4}	\\
ChatGPT 4o	&	unk.	&	88.3	&	71.5	&	73.1	&	84.5	\\
ChatGPT 4o mini	&	unk.	&	76.7	&	55.4	&	62.2	&	71.5	\\
Gemini 2.0 Flash	&	unk.	&	82.0	&	68.1	&	73.7	&	83.8	\\
LeChat	&	unk.	&	59.7	&	34.7	&	32.1	&	48.4	\\\hline

    \end{tabular}}
    \caption{Coarse-grained category results of language exams. \textbf{Bold} numbers indicate highest performance for each size category.}
    \label{tab:cat_coarse}
\end{table}

In order to answer this question, we analyse trends and patterns relating to the four categories mentioned in Section~\ref{sec:language_tests}. Table~\ref{tab:cat_coarse} contains the performances of the studied language models per category. 
From the results, we note that overall, LLMs perform almost consistently best in vocabulary questions while grammar appears to be the hardest category. 
Furthermore, although the Reading Comprehension and Conversation Comprehension questions are nominally different, there is no statistical difference between the performances of the LLMs in either category (according to a paired t-test).
However, when taking into account the size categories of the LLMs, \emph{large} models perform significantly better in the conversation comprehension category than reading comprehension questions which might be due to the tested LLMs being specifically instruction-tuned to hold conversations.

In addition to the coarse-grained ones, we created fine-grained categories for the questions in the language tests (see Appendix~\ref{sec:app_e} for more information on the categories). 
We present the results of the best-performing models in Table~\ref{tab:fg_cat}. 
Due to space issues, we put the entire table in Appendix~\ref{sec:app_d}.

The categorisation was done in two steps: first, we did an automatic categorisation in a supervised fashion using Claude 3.7 Sonnet and category explanations as well as a separate manually annotated language exam that were added to the prompt. 
We provide the descriptions for each category as well as the prompt for annotation in Appendices~\ref{sec:app_e} and \ref{sec:app_f}.
In a second step, we manually reviewed the automatically annotated questions to ensure correctness.
It is to note that for this categorisation, we do not differentiate between reading comprehension and conversation comprehension, and thusly group them together.

\begin{table*}[ht!]
    \centering
    \resizebox{\textwidth}{!}{
    \begin{tabular}{|l|r|r|r|r|r|r|r|r|}\hline
         model	&	Llama 3.1 405B	&	DeepSeek-R1 671B	&	ChatGPT 3.5	&	Claude 3.5 	&	ChatGPT 4o	&	ChatGPT 4o m	&	Gemini 2.0 &	LeChat	\\\hline	
V\_Idioms	    &	42.9	&	81.0	&	71.4	&	42.9	&	52.4	&	38.1	&	33.3	&	38.1	\\
V\_Noun	        &	67.9	&	94.0	&	91.7	&	85.7	&	86.9	&	79.8	&	81.0	&	65.5	\\
V\_Verb	        &	68.2	&	93.2	&	90.9	&	88.6	&	90.9	&	68.2	&	84.1	&	61.4	\\
V\_Adjective	&	47.8	&	91.3	&	100.0	&	100.0	&	95.7	&	87.0	&	87.0	&	47.8	\\
V\_Native	    &	33.3	&	100.0	&	50.0	&	50.0	&	66.7	&	16.7	&	16.7	&	50.0	\\
G\_Gender	    &	37.8	&	70.3	&	73.0	&	70.3	&	78.4	&	56.8	&	70.3	&	24.3	\\
G\_Pronouns	    &	40.0	&	65.0	&	60.0	&	75.0	&	60.0	&	65.0	&	70.0	&	50.0	\\
G\_Verb	        &	40.7	&	68.5	&	75.9	&	90.7	&	75.9	&	53.7	&	75.9	&	35.2	\\
G\_Adjective	&	22.2	&	55.6	&	44.4	&	66.7	&	55.6	&	22.2	&	66.7	&	44.4	\\
G\_Tense	    &	36.7	&	69.4	&	77.6	&	91.8	&	77.6	&	55.1	&	79.6	&	34.7	\\
G\_Past	        &	30.0	&	55.0	&	75.0	&	85.0	&	70.0	&	50.0	&	75.0	&	30.0	\\
G\_Present	    &	53.8	&	84.6	&	92.3	&	100.0	&	92.3	&	76.9	&	84.6	&	38.5	\\
G\_Subjunctive	&	28.6	&	71.4	&	71.4	&	92.9	&	78.6	&	42.9	&	71.4	&	35.7	\\
G\_Modal	    &	14.3	&	100.0	&	71.4	&	85.7	&	85.7	&	71.4	&	85.7	&	42.9	\\
G\_Prepositions	&	55.6	&	81.5	&	75.9	&	85.2	&	74.1	&	64.8	&	68.5	&	35.2	\\
G\_Cases	    &	33.3	&	64.1	&	69.2	&	69.2	&	64.1	&	51.3	&	53.8	&	35.9	\\
RC\_TF	        &	55.9	&	79.6	&	85.5	&	88.2	&	83.6	&	61.8	&	80.9	&	39.5	\\
RC\_Simple	    &	63.6	&	86.4	&	88.6	&	86.4	&	86.4	&	86.4	&	86.4	&	40.9	\\
RC\_Complex	    &	59.1	&	84.8	&	83.3	&	86.4	&	83.3	&	81.8	&	84.8	&	47.0	\\\hline

    \end{tabular}}
    \caption{Performance of large models on fine-grained categories. V: Vocabulary; G: Grammar; RC: Reading Comprehension; TF: True/False}
    \label{tab:fg_cat}
\end{table*}

For vocabulary-related questions, there is a clear pattern showing that LLMs perform worse in categories that require "language-exclusive" knowledge and cannot be easily inferred from other languages, i.e., idioms and native words.
Regarding grammar, the hardest categories for the large models are adjectives, past tense, and cases. 
The main challenge appears to be related to the fact that the possible answers for many of those questions are almost identical.
For example, one question from the adjectives category is:

{\small\textit{Op \_\_\_ Weiere kann een de Wanter Schlittschong fueren} }

\hspace{-14pt} with the possible answers being:

{\small\textit{zougefruer, zougefruerene, zougefruerenen, zougefréieren}}

For the reading comprehension questions, we split them into three categories: \emph{true/false}, \emph{simple}, and \emph{complex}. 
\textit{Simple} questions do not require an understanding of the entire given text excerpt, and the correct answer for those questions can be found either explicitly written in the text or by understanding a portion of the text.
\textit{True/false} and \textit{complex} questions require models to fully understand the given text excerpt, the main difference being that \textit{True/False} questions are binary while \textit{complex} ones have three or four possible answers. 
As such, the latter questions are considered harder. 
This assumption is indeed confirmed by the results as the \emph{large} models seem to perform best for \textit{simple} questions. 

It is to note, that these observations largely hold true for \emph{small} and \emph{medium-sized} models except for simple reading comprehension questions which was often the hardest category for those models. 

\subsection{RQ3: Are there correlations between performing well in a language exam and performing well in an NLG task?}
\begin{table}[ht!]
    \centering
    \begin{subtable}[b]{\linewidth}
    
    \resizebox{\textwidth}{!}{
    \begin{tabular}{|l|r|r|r|r|}
    \hline
Task	&	Small	&	Medium	&	Large	&	All	\\\hline
Claude-G	&	0.651	&	0.735	&	0.291	&	0.837	\\
Claude-A	&	0.648	&	0.585	&	0.301	&	0.794	\\
METEOR	&	0.427	&	0.450	&	0.089	&	0.667	\\
BERTScore	&	0.407	&	0.116	&	0.266	&	0.622	\\\hline

    \end{tabular}}
    \caption{Pearson Correlation Coefficients}
    \end{subtable}
    \begin{subtable}[b]{\linewidth}
    
    \resizebox{\textwidth}{!}{
    \begin{tabular}{|l|r|r|r|r|}
    \hline
		&	small	&	medium	&	large	&	all	\\\hline
Claude-G	&	6.127	&	7.743	&	2.171	&	10.938	\\
Claude-A	&	6.072	&	5.145	&	2.255	&	9.327	\\
METEOR	&	3.369	&	3.600	&	0.639	&	6.390	\\
BERTScore	&	3.179	&	0.837	&	1.972	&	5.676	\\\hline

    \end{tabular}}
    \caption{T-Scores}
    \end{subtable}
   \begin{subtable}[b]{\linewidth}
    
    \resizebox{\textwidth}{!}{
    \begin{tabular}{|l|r|r|r|r|}
    \hline
	&	small	&	medium	&	large	&	all	\\\hline
Claude-G	&	1.299E-07	&	3.667E-10	&	0.0346	&	5.552E-15	\\
Claude-A	&	1.583E-07	&	4.318E-06	&	0.0285	&	1.312E-12	\\
METEOR	&	0.0014	&	0.0007	&	0.5259	&	4.998E-08	\\
BERTScore	&	0.0025	&	0.4064	&	0.0541	&	6.578E-07	\\\hline

    \end{tabular}}
    \caption{P-Values}
    \end{subtable}
    \caption{Correlation tests between the overall exam scores and Headline Generation task performances.}
    \label{tab:pearson_headline}
\end{table}

\begin{figure*}[ht!]
\centering
\begin{subfigure}[]{.49\textwidth}

\centering
    	    \includegraphics[width=1.02\textwidth]{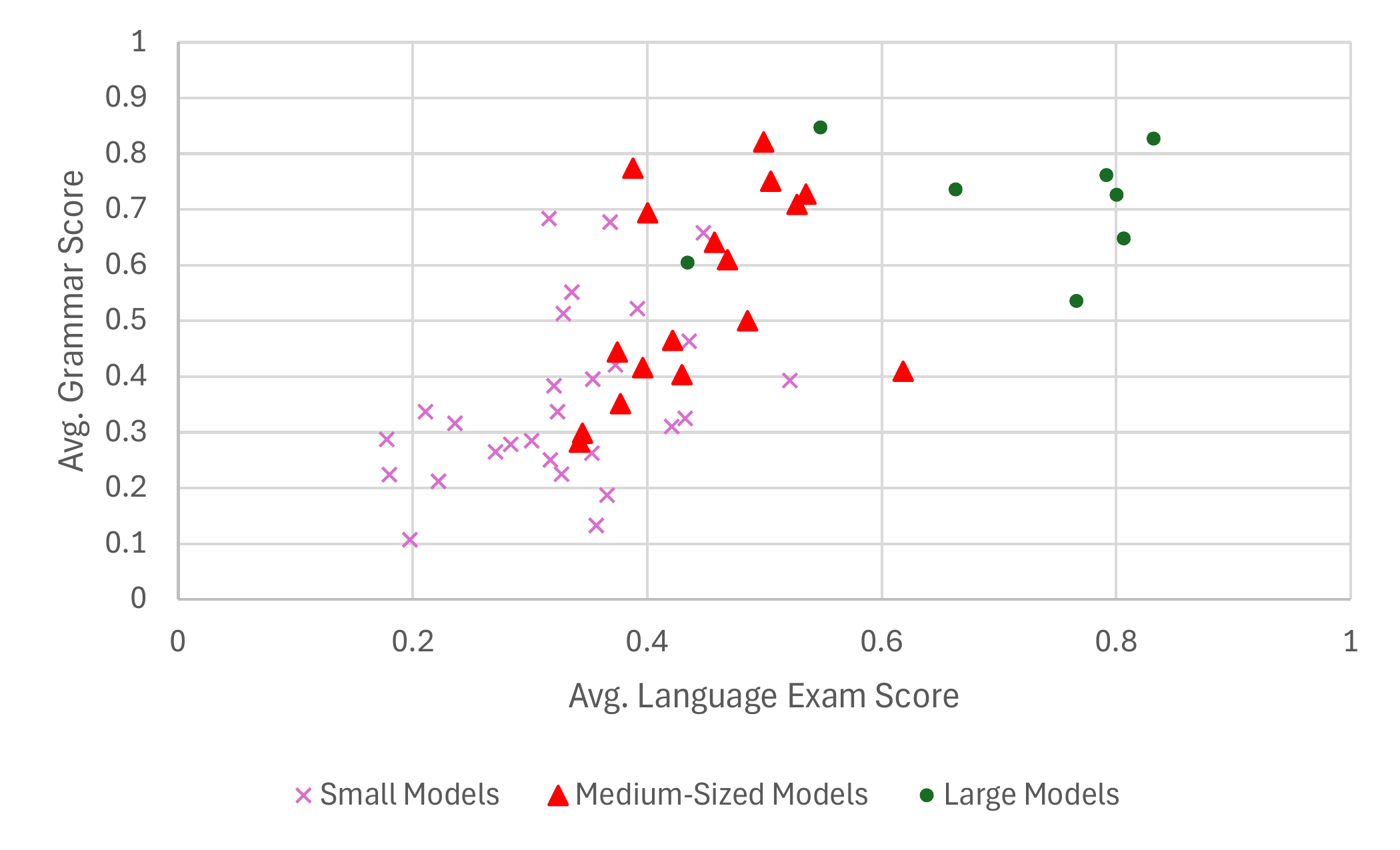}
            \caption{Claude Score - Grammar and Spelling}
            \label{fig:small_u}
\end{subfigure}
~
\begin{subfigure}[]{.49\textwidth}
\centering
            \includegraphics[width=1.02\textwidth]{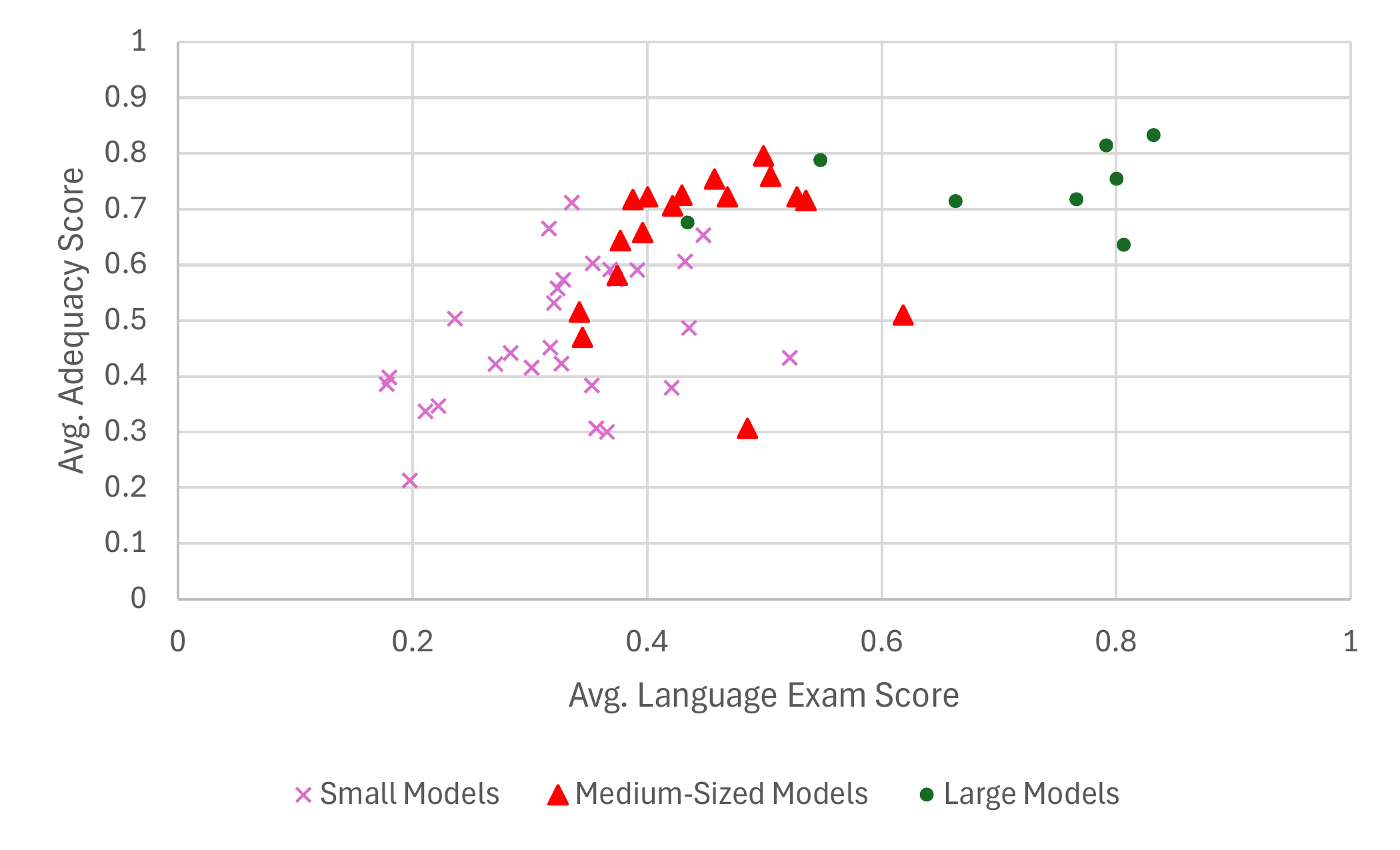}
            \caption{Claude Score - Adequacy}
            \label{fig:small_c}
\end{subfigure}
\centering
\begin{subfigure}[]{.49\textwidth}
\centering
    	    \includegraphics[width=1.02\textwidth]{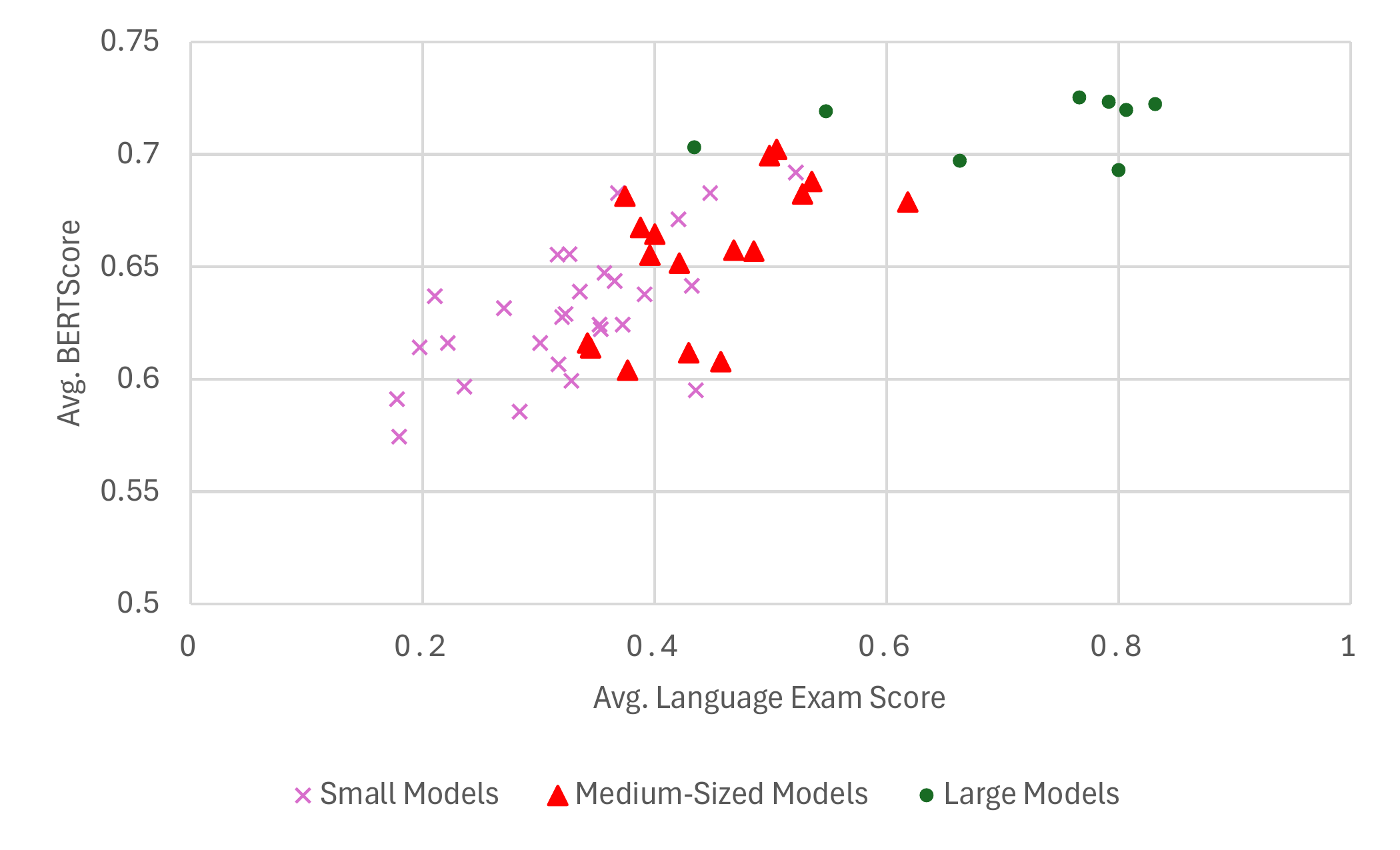}
            \caption{BERTScore}
            \label{fig:medium_u}
\end{subfigure}
~
\begin{subfigure}[]{.49\textwidth}
\centering
            \includegraphics[width=1.02\textwidth]{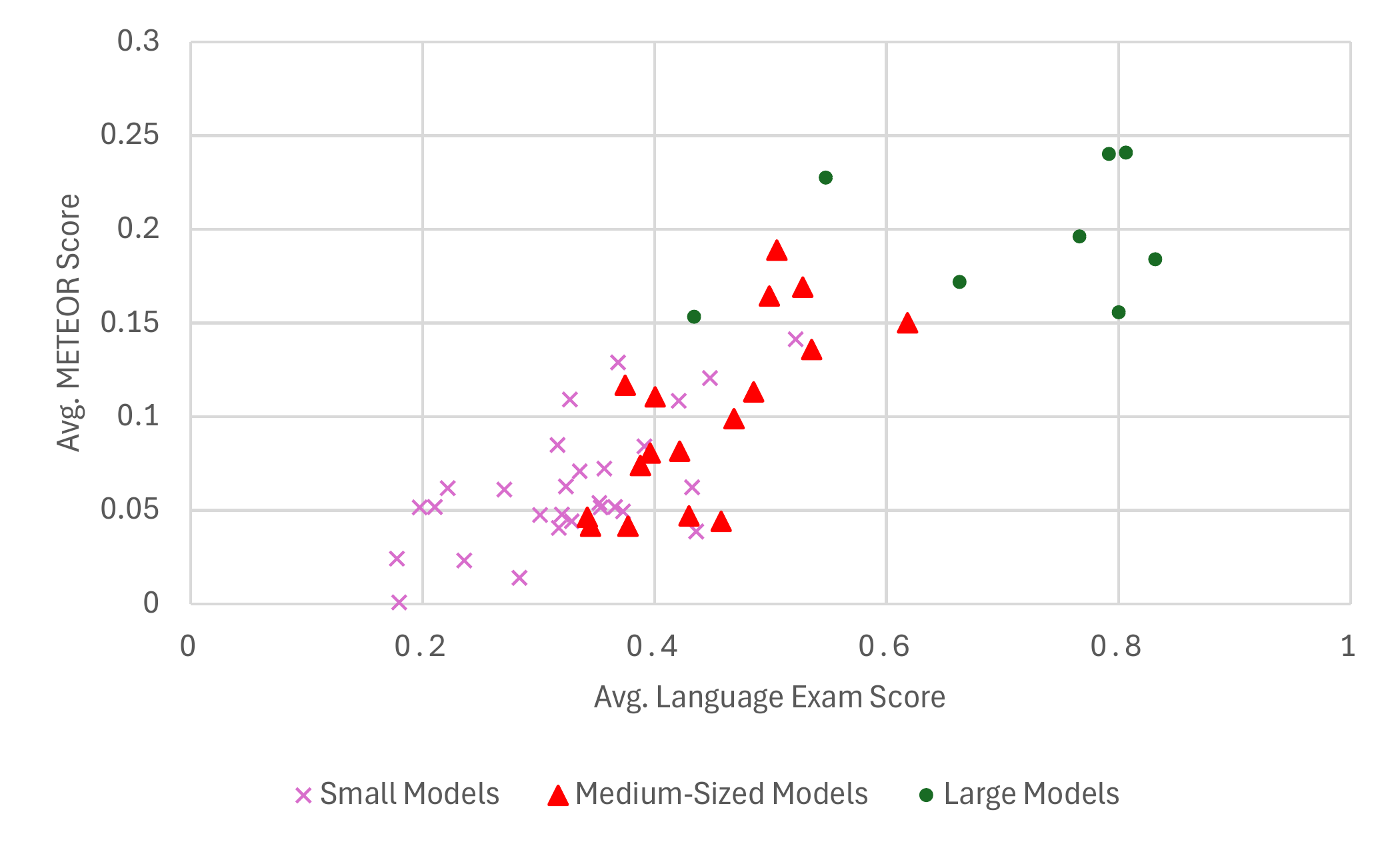}
            \caption{METEOR Score}
            \label{fig:medium_c}
\end{subfigure}
\caption{Performance on language exams vs performance on the Headline Generation task.}
\label{fig:luxgen}
\end{figure*}

For this research question, we first average out the performances of every LLM on all six language exams, resulting in their overall exam performance score.
We then let the LLMs solve both of the LuxGen tasks, and evaluate the outputs using an LLM-as-a-judge (LaaJ).
Our LLM of choice is Claude 3.7 Sonnet~\cite{claude2025}.
After manually verifying its capabilities to solve the tasks at hand, we deemed it an adequate LLM to evaluate the experiments as it solved both tasks well and with minimal spelling mistakes.
In addition to the LaaJ-metrics, we employ the METEOR and BERTScore metrics to measure the similarity between LLM outputs and the corresponding ground truths which also allows us to assess the usefulness of our novel metrics (see Appendix~\ref{sec:app_comp}).
For both tasks and all metrics we employ, we calculate mean performance scores for each LLM. 
Figure~\ref{fig:luxgen} shows the performances of all LLMs on the Headline Generation task ($y$-axis) plotted against their respective overall exam performances ($x$-axis). See Appendix~\ref{sec:app_desc} for plots for the Short Description task and Appendix~\ref{sec:app_g} for detailed results for both tasks.
Generally, the LuxGen performances of the LLMs appear to strongly positively correlate with their exam performances.
This is true for all metrics, although the spread and average performance is lower for the non-LaaJ metrics.

To confirm our observations, we calculate Pearson's Correlation Coefficients (PCCs) for each set of performances (see Table~\ref{tab:pearson_headline}). For most cases, we note moderate to strong PCCs. 
Some notable exceptions are the \emph{medium} and \emph{large} categories of LLMs with BERTScore and METEOR metrics, respectively, where there appears to be either a weak positive correlation or no correlation at all. We further calculate corresponding t-scores and p-values, and show that our findings are significant in most cases for $\alpha=0.05$ except for the aforementioned exceptions and BERTScore for the \emph{large} category.
We repeat the same significance tests for the Short Description task (see Appendix~\ref{sec:app_desc}) and find consistently high positive PCCs for all metrics, significant at $\alpha=0.05$.

%% file: 6_discussion.tex
\section{Discussion}
Generally, we conclude that language exams can indeed be a useful tool to assess the generation capabilities of LLMs. 
For the NLP tasks we tested, we can to a certain degree predict if an LLM performs well given how well it scored on the exams. 
However, there are some caveats to keep in mind:

Firstly, it is important to consider for what type of task an LLM is considered. 
Language exams are used to evaluate the general proficiency in a language across multiple disciplines. 
As such, the overall exam score can be dragged down by irrelevant categories of questions, e.g., an LLM used to detect hate speech in an online forum does not necessarily need to have a perfect grasp on obscure grammar rules. 
In such a case, it is preferable to adjust the exam to only include certain question categories.
Secondly, for this study, as there was no instruction tuning data available for Luxembourgish to the best of our knowledge, we did not do any additional fine-tuning which might have altered the performances of the tested LLMs to different degrees. 
It is possible that certain LLMs adapt better when fine-tuned on an appropriate instruction tuning dataset first.

%% file: 8_conclusion.tex
\section{Conclusion}

In this study, we investigated the performance of a large number of recent LLMs on official language exams, and assessed whether or not they serve as a useful tool to determine the capabilities of language models. 
We found that many models struggle with even the easier exams, with only the largest models scoring well in every exam.
We also delved into dividing the exam questions into fine-grained categories to assess which types of questions are harder to solve for LLMs, finding that questions requiring language-exclusive knowledge as well as questions with similar answers pose problems to many models.
Finally, we determined that using such exams can indeed be used as an indicator for language generation and understanding capabilities of LLMs, with some caveats to consider.

%% file: 09_acknowledgements.tex
\section{Acknowledgments}
This work is supported by the LLMs4EU project, funded by the European Union through the Digital Europe Programme (DIGITAL) under the grant agreement 101198470, and by the BESSER project through the Luxembourg National Research Fund (FNR) PEARL programme under the grant agreement 16544475.

We would like to thank the Institut National des Langues Luxembourg for providing the Luxembourgish exams for our study.

%% file: 10_limitations.tex
\section{Limitations}
While we are confident about the findings and conclusions of our studies, there are still certain limitations to consider. 
First, while we aimed to include a comprehensive list of current LLMs in our experiments, it is a nearly impossible task considering the frequent release of new models. We even had to scrap parts of our experiments that were done using Gemini 1.5 Flash as it became inaccessible after the release of the 2.0 model. We are also fully aware that many new model versions have been released that we did not consider for this study. We however remain confident that our selection of LLMs gives a good overview of available models.
A limitation that might have affected the performance on the proficiency exams is the availability of such exams in the training data. While the exams we received from INLL are not publicly available and we do not believe they could have been retrieved from a webcrawl or similar means, LLMs could have seen exams from other languages or simpler language skill placement tests in Luxembourgish. 
Another limitation relates to the relatively small number of generation tasks we considered for RQ3. As they are essentially both summarisation tasks, it is possible that our conclusions do not apply to other types of NLP tasks. This could be a useful research direction for future work.
In addition, we only used a small sample of those two summarisation tasks due to the large number of LLMs we considered.
Furthermore, as the data that was used to create those tasks are publicly available, it is possible that at least a subset of the studied LLMs saw that data during training which would have influenced their performance in those tasks.
Finally, it is possible that our conclusions are not generalisable to other languages since we are limiting ourselves to Luxembourgish. Again, studying if we can apply our methodology to other low-resource languages would be a worthwhile research direction that can benefit other language communities.

%% file: 10_appendix.tex
\newpage
\renewcommand{\thesubsection}{\Alph{subsection}}
\subsection{Number and Distribution of Open Questions in Language Exams}
\label{sec:app_pre_a}
\begin{table}[ht!]
    
    \centering
    \resizebox{.49\textwidth}{!}{
    \begin{tabular}{|r|l|l|l|l|l|l|l|}
    \hline
        Category & A1 & A2 & B1 & B2 & C1 & C2 & Total \\\hline
         Vocabulary &0 &0&13&13&14&12&52 \\
         Grammar &0 &0&0&17&14&25&56  \\\hline
         Total &0 &0&13&30&28&37&108\\\hline
    \end{tabular}}
    \caption{Number of open questions in language exams}
    \label{tab:stat_lang_tests}
\end{table}

\subsection{Full List of LLMs and Reasons for Inclusion}
\label{sec:app_a}
Table~\ref{tab:app_reasons} contains the full list of LLMs we include in this study.
\begin{table*}[]
    \centering
    \resizebox{\textwidth}{!}{
    \begin{tabular}{|l|l|r|}
    \hline
       Model	&	Size	&	Reason for Inclusion	\\\hline
\multicolumn{3}{|c|}{\textbf{Small Models} $(\leq 15B)$}\\\hline
Qwen 2	&	0.5	&	Supports German and Dutch	\\
Llama 3.2	&	1	&	Supports German	\\
Gemma 3 &   1  &    Officially supports only English but manages to generate Luxembourgish in experiments\\
Qwen 2	&	1.5	&	Supports German and Dutch	\\
StableLM 2	&	1.6	&	Supports German and Dutch	\\
EuroLLM &   1.7 &   Supports German and Dutch \\
Gemma 2	&	2	&	Was trained on multilingual data	\\
Llama 3.2	&	3	&	Supports German	\\
Phi 3	&	3.8	&	Tested on multilingual NLP tasks	\\
Phi 3.5	&	3.8	&	Tested on multilingual NLP tasks	\\
Gemma 3 &   4   &   Supports 140+ languages/ not explicitly supporting Luxembourgish \\
Qwen 2	&	7	&	Supports German and Dutch	\\
Llama 2	&	7	&	Was trained on multilingual data / not explicitly multilingual	\\
WizardLM 2	&	7	&	claims to be multilingual	\\
Mistral	&	7	&	European model / not explicitly multilingual	\\
Aya-23	&	8	&	Supports German and Dutch	\\
Llama 3.1	&	8	&	Supports German	\\
Llama 3	&	8	&	Performed well on multilingual NLP task including German	\\
DeepSeek-R1	&	8	&	smaller version of massively multilingual model	\\
GLM 4	&	9	&	Supports German	\\
EuroLLM	&	9	&	Supports German and Dutch	\\
Gemma 2	&	9	&	Training on multilingual data	\\
Mistral-Nemo	&	12	&	Supports German	\\
StableLM 2	&	12	&	Supports German and Dutch	\\
Gemma 3 &   12   &   Supports 140+ languages/ not explicitly supporting Luxembourgish \\
Llama 2	&	13	&	Was trained on multilingual data	\\
Phi 3	&	14	&	Tested on multilingual NLP tasks	\\
Phi 4 & 14 & Trained on German data\\\hline
\multicolumn{3}{|c|}{\textbf{Medium-sized Models} $(> 15B \& \leq 200B)$}\\\hline	
Mistral Small	&	22	&	Supports German and Dutch	\\
Gemma 2	&	27	&	Training on multilingual data	\\
Gemma 3 &   27   &   Supports 140+ languages/ not explicitly supporting Luxembourgish \\
QWQ &   32   &    Supports 29 languages/ not explicitly supporting Luxembourgish\\
Aya-23	&	35	&	Supports German and Dutch	\\
Command-R	&	35	&	Supports German	\\
Alfred	&	40	&	Supports German	\\
Mixtral	&	8x7	&	Supports German	\\
Llama 2	&	70	&	Was trained on multilingual data	\\
Llama 3	&	70	&	Performed well on multilingual NLP task including German	\\
Llama 3.1	&	70	&	Supports German	\\
DeepSeek-R1	&	70	&	smaller version of massively multilingual model	\\
Qwen 2	&	72	&	Supports German and Dutch	\\
Command-R+	&	104	&	Supports German	\\
Mistral-Large	&	123	&	Supports German	\\
Mixtral	&	8x22	&	Supports German	\\
WizardLM 2	&	8x22	&	claims to be multilingual	\\\hline
\multicolumn{3}{|c|}{\textbf{Large Models} $(> 200B)$}\\\hline

Llama 3.1	&	405	&	Supports German	\\
DeepSeek-R1	&	671	&	Massively Multilingual Model	\\
ChatGPT 3.5	&	175	&	Massively Multilingual Model	\\
Claude 3.5 Sonnet	&	unk	&	Massively Multilingual Model	\\
ChatGPT 4o	&	unk	&	Massively Multilingual Model	\\
ChatGPT 4o mini	&	unk	&	Massively Multilingual Model	\\
Gemini 2.0 Flash	&	unk	&	Massively Multilingual Model	\\
LeChat	&	unk	&	Massively Multilingual Model	\\\hline

    \end{tabular}}
    \caption{Full list of LLMs and our reasoning for including them in this study}
    \label{tab:app_reasons}
\end{table*}
\subsection{Language Exam Prompt}
\label{sec:app_a1}
We use the following prompt for testing LLMs on language exams: 
\begin{verbatim}
    
"I will give you a language test for 
Luxembourgish. For each part, you will 
get a TEXT with one part missing marked 
by [BLANK] and a list of possible 
ANSWERS where every possible option is 
separated with a comma. You MUST choose 
the option out of the provided list that 
best fits as replacement for the given 
BLANK. As output, ONLY write the chosen 
option, nothing else. DO NOT WRITE ANY 
ADDITIONAL TEXT."

INPUT: [the question and possible answers]
OUTPUT:
    
\end{verbatim}
where the INPUTs for the exams are formatted as follows:
\begin{verbatim}
TEXT: Den Direkter ass houfereg [BLANK] 
eis Leeschtung.	
ANSWERS: [iwwert,fir,mat,op]
\end{verbatim}

\subsection{Evaluation Metric Prompts for LuxGen}
\label{sec:app_b}
For each task and metric, we designed different prompts for a total of four prompts:

\begin{itemize}[leftmargin=0pt]
    \item[] \textbf{Grammar and Spelling Metric for Small Description Task:}
\begin{verbatim}
I will give you a table as a TSV file. This 
table contains Luxembourgish Wikipedia 
articles in the first column, the second 
column contains ground truth short 
descriptions for the articles. The rest of 
the columns contain short descriptions 
generated by various language models. The 
generated descriptions are supposed to be 
written in correct Luxembourgish. Your 
task is to judge the generated descriptions 
in terms of grammatical correctness and 
spelling. Keep in mind that the descriptions 
do not need to be complete sentences. 

For each language model and sample, return 
a score ranging from 0 to 1 with two 
decimals behind the comma. For the score, 
1 means that the description is perfectly 
written without any grammatical or 
spelling mistakes. The more mistakes there 
are, the lower the score should be. A 0 
means that the headline is full of mistakes 
and made up words. Also return 0 if the 
description is not written in 
Luxembourgish at all.

Return the average scores in a table.

\end{verbatim}

    \item[] \textbf{Adequacy Metric for Small Description Task:}
\begin{verbatim}

I will give you a table as a TSV file. This 
table contains Luxembourgish Wikipedia 
articles in the first column, the second 
column contains ground truth short 
descriptions for the articles. The rest of 
the columns contain short descriptions 
generated by various language models. The 
generated descriptions are supposed to be 
written in correct Luxembourgish. Your 
task is to judge the generated 
descriptions in terms of the content 
being an adequate description of the 
article. Keep in mind that the 
descriptions do not need to be complete 
sentences. 

For each language model and sample, return 
a score ranging from 0 to 1 with two 
decimals behind the comma. For the score, 
1 means that the description is identical 
to the ground truth, and should be lower 
the more it deviates from the meaning of 
the article. The score should also be 
lowered for descriptions that are too long.
A 0 means that the description is 
completely irrelevant to the article.

Return the average scores in a table.

\end{verbatim}

    \item[] \textbf{Grammar and Spelling Metric for News Headline Task:}
\begin{verbatim}
You are a news editor for a Luxembourgish 
news paper. I will give you a table as a 
TSV file. This table contains news 
articles in the first column, the second 
column contains ground truth headlines 
for the articles. The rest of the columns 
will contain headlines that were produced 
by various language models. Your task is 
to judge the produced headlines in terms 
of grammatical correctness and spelling. 

For each language model and sample, return 
a score ranging from 0 to 1 with two 
decimals behind the comma. For the score, 
1 means that the headline is perfectly 
written without any grammatical or 
spelling mistakes. The more mistakes there 
are, the lower the score should be. A 0 
means that the headline is full of 
mistakes and made up words. Also return 
0 if the description is not written in 
Luxembourgish at all. 

Return the average scores in a table.

\end{verbatim}

    \item[] \textbf{Adequacy Metric for News Headline Task:}
\begin{verbatim}
You are a news editor for a Luxembourgish 
news paper. I will give you a table as a 
TSV file. This table contains news 
articles in the first column, the second 
column contains ground truth headlines 
for the articles. The rest of the columns 
will contain headlines that were produced 
by various language models. Your task is 
to judge the produced headlines in terms 
of summary of the content of the article. 

For each language model and sample, 
return a score ranging from 0 to 1 with 
two decimals behind the comma. For the 
score, a 1 means that headline is 
identical to the ground truth, and should 
decrease the more its meaning deviates 
from the content in the news article. The 
score should also be lowered for headlines 
that are too long. A 0 means that the 
headline is completely irrelevant to the 
article.

Return the average scores in a table.

\end{verbatim}

\end{itemize}

\subsection{Coarse-grained Exam Results}
\label{sec:app_c}
Table~\ref{tab:app_detail_coarse} shows the full breakdown of our results on language tests for each language level.

\subsection{Fine-grained Exam Results}
\label{sec:app_d}
Table~\ref{tab:app_detail_fine} shows results of our experiments on fine-grained categories for all LLMs.

\subsection{Fine-grained Category Description}
\label{sec:app_e}
Table~\ref{tab:app_cat_fine} contains explanations and statistics for the fine-grained categories we created for the exam dataset.

\subsection{Fine-grained Category Annotation Prompt}

\label{sec:app_f}
We use the following prompt to categorise the exam questions:

\begin{verbatim}
I have a language test for Luxembourgish 
where I would like you to categorise every 
question into different types of skill 
they are testing. For every question and 
every category, determine and mark 
whether or not a given question belongs 
to a given category, as follows: 

0, the question does not belong 
to the category 
1, the question belongs to the category 

Write the output in tabular form like 
TSV.

The file I am uploading are some 
questions that I already manually 
annotated. 
Use it as a guideline. 
The list of categories and explanations 
is as follows:
vocabulary: any question relating 
to vocabulary 
grammar: any question relating 
to grammar 
vocabulary_idiom: a question relating 
to idioms or metaphors 
vocabulary_noun: a vocabulary question 
relating to a noun
vocabulary_verb: a vocabulary question 
relating to a verb
vocabulary_adjective: a vocabulary 
question relating to an adjective
vocabulary_native_word: a vocabulary 
question where the correct answer is a 
word exclusive to Luxembourgish 
grammar_gender: a question relating to determining the gender
of a determinant, adjective, or pronoun 
grammar_pronouns: a question relating to pronouns
grammar_verbs:  a question relating to conjugations of verbs
grammar_adjectives:  a question relating to adjectives
grammar_tense: a question relating to the correct use of tenses 
grammar_tense_past: a question relating to the past tense
grammar_tense_present: a question relating to the present tense 
grammar_tense_subjunctive: a question relating to the subjunctive tense
grammar_verbs_modal_verbs: a question relating to modal verbs 
grammar_preposition_conjunction: a question relating to the correct use 
of prepositions or conjunctions 
grammar_cases: a question relating to nominative,genitive,dative, or accusative
reading_comprehension_true_false: a reading comprehension question where the answer 
is either true or false
reading_comprehension_simple: a reading comprehension question that does not require to 
understand the entire text excerpt to answer correctly
reading_comprehension_complex: a reading comprehension question that requires a deeper 
understanding of the entire text excerpt to answer correctly
\end{verbatim}
\subsection{LuxGen Evaluation Results using LLM-as-a-Judge metrics}
\label{sec:app_g}
Table~\ref{tab:app_luxgen} shows the full evaluation results of the LuxGen experiments over three runs as well as the average performances.
\clearpage
\begin{table*}[]
    \centering
        \rotatebox{90}{

    \resizebox{1.1\textwidth}{!}{
    \begin{tabular}{|l|l||r|r|r|r||r|r|r|r||r|r|r|r||r|r|r|r||r|r|r|r||r|r|r|r|}
    \hline
	&&		\multicolumn{4}{|c||}{ \textbf{A1}}&									\multicolumn{4}{c||}{ \textbf{A2}}&								\multicolumn{4}{c||}{ \textbf{B1}}&								\multicolumn{4}{c||}{ \textbf{B2}}&								\multicolumn{4}{c||}{ \textbf{C1}}&								\multicolumn{4}{c|}{ \textbf{C2}}\\\hline							
Model	&	Size	&	Voc.	&	Gra.	&	RC	&	CC	&	Voc.	&	Gra.	&	RC	&	CC	&	Voc.	&	Gra.	&	RC	&	CC	&	Voc.	&	Gra.	&	RC	&	CC	&	Voc.	&	Gra.	&	RC	&	CC	&	Voc.	&	Gra.	&	RC	&	CC	\\\hline
																																				
Qwen 2	&	0.5	&	38.5	&	26.9	&	23.1	&	19.2	&	34.6	&	11.5	&	19.2	&	23.1	&	32.0	&	15.4	&	23.1	&	23.1	&	25.0	&	17.6	&	19.2	&	30.8	&	16.7	&	20.7	&	30.8	&	20.0	&	7.7	&	26.1	&	11.5	&	19.2	\\
Llama 3.2	&	1	&	15.4	&	34.6	&	34.6	&	19.2	&	26.9	&	11.5	&	19.2	&	19.2	&	32.0	&	46.2	&	19.2	&	15.4	&	21.4	&	20.6	&	19.2	&	23.1	&	16.7	&	20.7	&	11.5	&	28.0	&	15.4	&	8.7	&	19.2	&	7.7	\\
Gemma 3	&	1	&	38.5	&	46.2	&	50.0	&	46.2	&	19.2	&	34.6	&	50.0	&	42.3	&	44.0	&	42.3	&	34.6	&	34.6	&	14.3	&	44.1	&	50.0	&	53.8	&	29.2	&	20.7	&	26.9	&	24.0	&	19.2	&	39.1	&	38.5	&	38.5	\\
Qwen 2	&	1.5	&	34.6	&	38.5	&	42.3	&	15.4	&	38.5	&	26.9	&	23.1	&	23.1	&	28.0	&	30.8	&	19.2	&	23.1	&	28.6	&	38.2	&	23.1	&	34.6	&	25.0	&	13.8	&	19.2	&	28.0	&	23.1	&	21.7	&	23.1	&	26.9	\\
StableLM 2	&	1.6	&	30.8	&	34.6	&	7.7	&	15.4	&	19.2	&	23.1	&	0.0	&	15.4	&	20.0	&	30.8	&	11.5	&	23.1	&	25.0	&	23.5	&	11.5	&	3.8	&	20.8	&	24.1	&	3.8	&	12.0	&	23.1	&	34.8	&	0.0	&	11.5	\\
EuroLLM	&	1.7	&	19.2	&	15.4	&	11.5	&	23.1	&	23.1	&	23.1	&	15.4	&	23.1	&	24.0	&	11.5	&	11.5	&	3.8	&	21.4	&	29.4	&	23.1	&	19.2	&	33.3	&	27.6	&	15.4	&	28.0	&	23.1	&	13.0	&	23.1	&	11.5	\\
Gemma 2	&	2	&	38.5	&	46.2	&	38.5	&	34.6	&	38.5	&	23.1	&	34.6	&	34.6	&	44.0	&	19.2	&	30.8	&	30.8	&	17.9	&	32.4	&	30.8	&	50.0	&	29.2	&	31.0	&	19.2	&	32.0	&	38.5	&	26.1	&	23.1	&	26.9	\\
Llama 3.2	&	3	&	34.6	&	34.6	&	46.2	&	34.6	&	38.5	&	23.1	&	26.9	&	53.8	&	32.0	&	26.9	&	23.1	&	23.1	&	32.1	&	26.5	&	42.3	&	42.3	&	37.5	&	24.1	&	23.1	&	40.0	&	15.4	&	34.8	&	30.8	&	19.2	\\
Phi 3	&	3.8	&	26.9	&	46.2	&	57.7	&	50.0	&	34.6	&	23.1	&	38.5	&	50.0	&	24.0	&	23.1	&	23.1	&	38.5	&	25.0	&	26.5	&	15.4	&	38.5	&	33.3	&	24.1	&	23.1	&	32.0	&	23.1	&	8.7	&	19.2	&	19.2	\\
Phi 3.5	&	3.8	&	34.6	&	50.0	&	30.8	&	42.3	&	26.9	&	30.8	&	15.4	&	38.5	&	8.0	&	19.2	&	11.5	&	7.7	&	14.3	&	8.8	&	11.5	&	7.7	&	12.5	&	6.9	&	7.7	&	4.0	&	7.7	&	4.3	&	15.4	&	15.4	\\
Gemma 3	&	4	&	65.4	&	26.9	&	69.2	&	57.7	&	38.5	&	50.0	&	34.6	&	50.0	&	48.0	&	34.6	&	30.8	&	46.2	&	57.1	&	29.4	&	42.3	&	61.5	&	33.3	&	17.2	&	53.8	&	44.0	&	34.6	&	39.1	&	30.8	&	23.1	\\
Qwen 2	&	7	&	42.3	&	26.9	&	69.2	&	53.8	&	23.1	&	38.5	&	38.5	&	46.2	&	28.0	&	23.1	&	15.4	&	61.5	&	28.6	&	35.3	&	38.5	&	42.3	&	58.3	&	24.1	&	38.5	&	40.0	&	34.6	&	39.1	&	15.4	&	38.5	\\
Llama 2	&	7	&	26.9	&	34.6	&	53.8	&	30.8	&	46.2	&	23.1	&	34.6	&	50.0	&	24.0	&	26.9	&	26.9	&	19.2	&	46.4	&	32.4	&	38.5	&	42.3	&	29.2	&	17.2	&	19.2	&	36.0	&	11.5	&	26.1	&	38.5	&	30.8	\\
WizardLM 2	&	7	&	42.3	&	34.6	&	65.4	&	34.6	&	50.0	&	34.6	&	26.9	&	42.3	&	24.0	&	15.4	&	15.4	&	30.8	&	42.9	&	23.5	&	30.8	&	30.8	&	29.2	&	37.9	&	23.1	&	24.0	&	26.9	&	47.8	&	11.5	&	34.6	\\
Mistral	&	7	&	30.8	&	26.9	&	46.2	&	38.5	&	23.1	&	30.8	&	42.3	&	46.2	&	40.0	&	34.6	&	26.9	&	34.6	&	28.6	&	23.5	&	26.9	&	46.2	&	41.7	&	27.6	&	34.6	&	36.0	&	23.1	&	26.1	&	30.8	&	26.9	\\
Aya-23	&	8	&	38.5	&	38.5	&	46.2	&	38.5	&	30.8	&	19.2	&	30.8	&	42.3	&	36.0	&	19.2	&	26.9	&	38.5	&	25.0	&	32.4	&	26.9	&	42.3	&	29.2	&	41.4	&	23.1	&	32.0	&	30.8	&	26.1	&	38.5	&	30.8	\\
Llama 3.1	&	8	&	50.0	&	61.5	&	57.7	&	57.7	&	30.8	&	30.8	&	26.9	&	46.2	&	16.0	&	38.5	&	19.2	&	57.7	&	32.1	&	23.5	&	53.8	&	26.9	&	20.8	&	20.7	&	50.0	&	32.0	&	30.8	&	26.1	&	30.8	&	46.2	\\
Llama 3	&	8	&	42.3	&	42.3	&	50.0	&	46.2	&	30.8	&	34.6	&	26.9	&	50.0	&	40.0	&	30.8	&	30.8	&	38.5	&	25.0	&	38.2	&	34.6	&	46.2	&	45.8	&	17.2	&	26.9	&	28.0	&	23.1	&	39.1	&	42.3	&	23.1	\\
DeepSeek-R1	&	8	&	69.2	&	42.3	&	57.7	&	46.2	&	46.2	&	38.5	&	26.9	&	46.2	&	48.0	&	42.3	&	30.8	&	34.6	&	35.7	&	38.2	&	46.2	&	38.5	&	29.2	&	24.1	&	50.0	&	36.0	&	15.4	&	21.7	&	30.8	&	46.2	\\
GLM 4	&	9	&	57.7	&	42.3	&	69.2	&	50.0	&	34.6	&	19.2	&	42.3	&	69.2	&	56.0	&	15.4	&	26.9	&	53.8	&	60.7	&	26.5	&	50.0	&	50.0	&	54.2	&	34.5	&	46.2	&	44.0	&	38.5	&	26.1	&	26.9	&	50.0	\\
EuroLLM	&	9	&	53.8	&	30.8	&	57.7	&	34.6	&	50.0	&	38.5	&	42.3	&	50.0	&	32.0	&	42.3	&	30.8	&	26.9	&	39.3	&	23.5	&	34.6	&	42.3	&	41.7	&	31.0	&	23.1	&	32.0	&	23.1	&	26.1	&	26.9	&	26.9	\\
Gemma 2	&	9	&	76.9	&	50.0	&	61.5	&	46.2	&	61.5	&	38.5	&	30.8	&	57.7	&	48.0	&	26.9	&	34.6	&	46.2	&	64.3	&	38.2	&	53.8	&	42.3	&	58.3	&	20.7	&	46.2	&	40.0	&	34.6	&	30.4	&	42.3	&	30.8	\\
Mistral-Nemo	&	12	&	61.5	&	50.0	&	53.8	&	46.2	&	34.6	&	26.9	&	38.5	&	46.2	&	36.0	&	15.4	&	23.1	&	38.5	&	28.6	&	23.5	&	38.5	&	46.2	&	37.5	&	24.1	&	26.9	&	32.0	&	19.2	&	26.1	&	26.9	&	11.5	\\
StableLM 2	&	12	&	42.3	&	15.4	&	30.8	&	23.1	&	4.62	&19.2	&34.6	&42.3	&	44.0& 11.5	&	19.2	&	30.8	&	21.4	&	17.6	&	50.0	&	30.8	&	37.5	&	17.2	&	19.2	&	40.0	&	19.2	&	21.7	&	30.8	&	23.1	\\
Gemma 3	&	12	&	76.9	&	30.8	&	73.1	&	69.2	&	80.8	&	42.3	&	46.2	&	65.4	&	64.0	&	46.2	&	42.3	&	61.5	&	64.3	&	29.4	&	57.7	&	57.7	&	54.2	&	24.1	&	61.5	&	52.0	&	38.5	&	21.7	&	46.2	&	53.8	\\
Llama 2	&	13	&	34.6	&	38.5	&	15.4	&	23.1	&	19.2	&	11.5	&	15.4	&	11.5	&	8.0	&	34.6	&	19.2	&	30.8	&	32.1	&	20.6	&	30.8	&	23.1	&	25.0	&	24.1	&	23.1	&	20.0	&	34.6	&	30.4	&	23.1	&	19.2	\\
Phi 3	&	14	&	42.3	&	42.3	&	50.0	&	46.2	&	38.5	&	42.3	&	30.8	&	42.3	&	40.0	&	38.5	&	30.8	&	46.2	&	46.4	&	35.3	&	34.6	&	42.3	&	25.0	&	34.5	&	30.8	&	20.0	&	11.5	&	21.7	&	19.2	&	34.6	\\
Phi 4	&	14	&	50.0	&	50.0	&	65.4	&	50.0	&	65.4	&	42.3	&	50.0	&	50.0	&	36.0	&	42.3	&	34.6	&	57.7	&	57.1	&	44.1	&	46.2	&	30.8	&	41.7	&	24.1	&	34.6	&	40.0	&	26.9	&	17.4	&	42.3	&	46.2	\\
\hline																																																			
&&\multicolumn{24}{c|}{\textbf{Medium-sized Models} $(> 15B \& \leq 200B)$}\\\hline																																																			
																																																			
Mistral Small	&	22	&	46.2	&	34.6	&	69.2	&	57.7	&	57.7	&	19.2	&	34.6	&	73.1	&	44.0	&	19.2	&	15.4	&	46.2	&	42.9	&	38.2	&	30.8	&	46.2	&	45.8	&	31.0	&	38.5	&	40.0	&	11.5	&	34.8	&	19.2	&	38.5	\\
Gemma 2	&	27	&	65.4	&	46.2	&	80.8	&	73.1	&	73.1	&	50.0	&	57.7	&	73.1	&	68.0	&	38.5	&	42.3	&	69.2	&	60.7	&	26.5	&	57.7	&	42.3	&	62.5	&	24.1	&	50.0	&	68.0	&	30.8	&	34.8	&	53.8	&	46.2	\\
Gemma 3	&	27	&	76.9	&	46.2	&	84.6	&	76.9	&	96.2	&	57.7	&	61.5	&	73.1	&	76.0	&	61.5	&	42.3	&	69.2	&	67.9	&	52.9	&	65.4	&	57.7	&	75.0	&	48.3	&	61.5	&	60.0	&	34.6	&	43.5	&	50.0	&	50.0	\\
QWQ	&	32	&	80.8	&	46.2	&	61.5	&	53.8	&	61.5	&	42.3	&	30.8	&	61.5	&	64.0	&	42.3	&	34.6	&	65.4	&	50.0	&	38.2	&	53.8	&	50.0	&	62.5	&	20.7	&	34.6	&	40.0	&	38.5	&	39.1	&	34.6	&	65.4	\\
Aya-23	&	35	&	34.6	&	34.6	&	61.5	&	42.3	&	46.2	&	15.4	&	38.5	&	50.0	&	40.0	&	46.2	&	30.8	&	42.3	&	57.1	&	17.6	&	46.2	&	26.9	&	25.0	&	31.0	&	34.6	&	44.0	&	34.6	&	39.1	&	30.8	&	34.6	\\
Command-R	&	35	&	42.3	&	53.8	&	61.5	&	46.2	&	46.2	&	19.2	&	38.5	&	50.0	&	36.0	&	26.9	&	30.8	&	38.5	&	35.7	&	35.3	&	50.0	&	50.0	&	37.5	&	27.6	&	38.5	&	40.0	&	38.5	&	43.5	&	46.2	&	23.1	\\
Alfred	&	40	&	30.8	&	38.5	&	61.5	&	38.5	&	50.0	&	30.8	&	26.9	&	50.0	&	24.0	&	30.8	&	26.9	&	34.6	&	28.6	&	32.4	&	30.8	&	26.9	&	45.8	&	31.0	&	34.6	&	36.0	&	26.9	&	30.4	&	26.9	&	34.6	\\
Mixtral	&	56	&	50.0	&	38.5	&	61.5	&	50.0	&	50.0	&	19.2	&	42.3	&	53.8	&	48.0	&	19.2	&	26.9	&	42.3	&	50.0	&	38.2	&	34.6	&	34.6	&	50.0	&	13.8	&	23.1	&	44.0	&	34.6	&	34.8	&	30.8	&	19.2	\\
Llama 2	&	70	&	23.1	&	34.6	&	57.7	&	38.5	&	34.6	&	26.9	&	30.8	&	42.3	&	24.0	&	26.9	&	26.9	&	34.6	&	39.3	&	35.3	&	30.8	&	53.8	&	25.0	&	24.1	&	26.9	&	28.0	&	26.9	&	21.7	&	46.2	&	61.5	\\
Llama 3	&	70	&	61.5	&	34.6	&	65.4	&	50.0	&	73.1	&	30.8	&	38.5	&	53.8	&	36.0	&	38.5	&	38.5	&	42.3	&	42.9	&	26.5	&	38.5	&	34.6	&	58.3	&	31.0	&	34.6	&	32.0	&	46.2	&	52.2	&	26.9	&	50.0	\\
Llama 3.1	&	70	&	57.7	&	38.5	&	69.2	&	57.7	&	84.6	&	34.6	&	42.3	&	65.4	&	56.0	&	53.8	&	34.6	&	53.8	&	71.4	&	44.1	&	46.2	&	50.0	&	58.3	&	48.3	&	42.3	&	40.0	&	34.6	&	56.5	&	38.5	&	38.5	\\
DeepSeek-R1	&	70	&	76.9	&	53.8	&	65.4	&	61.5	&	69.2	&	30.8	&	30.8	&	73.1	&	68.0	&	61.5	&	42.3	&	53.8	&	67.9	&	35.3	&	50.0	&	69.2	&	62.5	&	41.4	&	38.5	&	56.0	&	42.3	&	30.4	&	42.3	&	50.0	\\
Qwen 2	&	72	&	65.4	&	53.8	&	65.4	&	57.7	&	50.0	&	34.6	&	42.3	&	57.7	&	44.0	&	34.6	&	38.5	&	46.2	&	57.1	&	41.2	&	57.7	&	53.8	&	62.5	&	24.1	&	42.3	&	40.0	&	30.8	&	30.4	&	46.2	&	53.8	\\
Command-R+	&	104	&	57.7	&	38.5	&	61.5	&	53.8	&	50.0	&	19.2	&	42.3	&	42.3	&	52.0	&	30.8	&	30.8	&	46.2	&	42.9	&	47.1	&	42.3	&	46.2	&	45.8	&	20.7	&	34.6	&	28.0	&	30.8	&	30.4	&	30.8	&	38.5	\\
Mistral-Large	&	123	&	84.6	&	57.7	&	73.1	&	50.0	&	65.4	&	34.6	&	38.5	&	61.5	&	44.0	&	42.3	&	30.8	&	53.8	&	75.0	&	38.2	&	50.0	&	50.0	&	66.7	&	34.5	&	26.9	&	48.0	&	53.8	&	39.1	&	34.6	&	50.0	\\
Mixtral	&	176	&	57.7	&	57.7	&	73.1	&	61.5	&	57.7	&	50.0	&	38.5	&	50.0	&	44.0	&	30.8	&	38.5	&	53.8	&	60.7	&	35.3	&	34.6	&	38.5	&	41.7	&	34.5	&	38.5	&	44.0	&	34.6	&	34.8	&	30.8	&	57.7	\\
WizardLM 2	&	176	&	53.8	&	50.0	&	65.4	&	65.4	&	42.3	&	30.8	&	30.8	&	57.7	&	32.0	&	38.5	&	38.5	&	53.8	&	50.0	&	41.2	&	46.2	&	34.6	&	37.5	&	31.0	&	30.8	&	44.0	&	26.9	&	34.8	&	38.5	&	38.5	\\\hline
&&\multicolumn{24}{c|}{\textbf{Large Models} $(> 200B)$}\\\hline																																																			
Llama 3.1	&	405	&	92.3	&	57.7	&	80.8	&	84.6	&	26.9	&	38.5	&	34.6	&	57.7	&	72.0	&	46.2	&	26.9	&	50.0	&	67.9	&	47.1	&	65.4	&	61.5	&	79.2	&	37.9	&	57.7	&	52.0	&	42.3	&	30.4	&	61.5	&	50.0	\\
DeepSeek-R1	&	671	&	96.2	&	84.6	&	96.2	&	92.3	&	100.0	&	65.4	&	76.9	&	92.3	&	88.0	&	76.9	&	42.3	&	88.5	&	89.3	&	58.8	&	84.6	&	76.9	&	83.3	&	65.5	&	73.1	&	72.0	&	96.2	&	100.0	&	73.1	&	57.7	\\
ChatGPT 3.5	&	unk	&	100.0	&	76.9	&	100.0	&	92.3	&	100.0	&	84.6	&	69.2	&	88.5	&	96.0	&	92.3	&	38.5	&	96.2	&	89.3	&	76.5	&	80.8	&	88.5	&	87.5	&	58.6	&	76.9	&	84.0	&	76.9	&	43.5	&	76.9	&	65.4	\\
Claude 3.5 S	&	unk	&	100.0	&	80.8	&	100.0	&	88.5	&	96.2	&	88.5	&	88.5	&	96.2	&	96.0	&	92.3	&	38.5	&	96.2	&	85.7	&	82.4	&	84.6	&	96.2	&	87.5	&	72.4	&	80.8	&	72.0	&	61.5	&	78.3	&	69.2	&	69.2	\\
ChatGPT 4o	&	unk	&	96.2	&	84.6	&	100.0	&	88.5	&	100.0	&	73.1	&	80.8	&	88.5	&	96.0	&	84.6	&	38.5	&	96.2	&	92.9	&	79.4	&	80.8	&	84.6	&	83.3	&	55.2	&	73.1	&	88.0	&	61.5	&	52.2	&	65.4	&	61.5	\\
ChatGPT 4o m	&	unk	&	96.2	&	73.1	&	100.0	&	92.3	&	100.0	&	84.6	&	80.8	&	96.2	&	68.0	&	46.2	&	42.3	&	80.8	&	75.0	&	55.9	&	50.0	&	57.7	&	75.0	&	37.9	&	57.7	&	52.0	&	46.2	&	34.8	&	42.3	&	50.0	\\
Gemini 2.0 F	&	unk	&	88.5	&	76.9	&	96.2	&	96.2	&	96.2	&	65.4	&	80.8	&	96.2	&	88.0	&	84.6	&	42.3	&	84.6	&	89.3	&	67.6	&	80.8	&	88.5	&	87.5	&	62.1	&	80.8	&	76.0	&	42.3	&	52.2	&	61.5	&	61.5	\\
LeChat	&	unk	&	80.8	&	26.9	&	3.8	&	57.7	&	73.1	&	50.0	&	38.5	&	53.8	&	44.0	&	26.9	&	34.6	&	50.0	&	71.4	&	32.4	&	53.8	&	53.8	&	54.2	&	41.4	&	26.9	&	44.0	&	34.6	&	30.4	&	34.6	&	30.8	\\\hline
    \end{tabular}}}
    \caption{Detailed results on experiments with coarse-grained categories}
    \label{tab:app_detail_coarse}
\end{table*}
\clearpage
\begin{table*}[h!]
    \centering
    \rotatebox{90}{
    \resizebox{1.34\textwidth}{!}{
    \begin{tabular}{|l|l|r|r|r|r|r|r|r|r|r|r|r|r|r|r|r|r|r|r|r|}
    \hline
Model	&	Size	&	V\_Idioms	&	V\_Noun	&	V\_Verb	&	V\_Adj.	&	V\_Native	&	G\_Gender	&	G\_Pron.	&	G\_Verb	&	G\_Adj.	&	G\_Tense	&	G\_Past	&	G\_Present	&	G\_Subj.	&	G\_Modal	&	G\_Prep. 	&	G\_Cases	&	RC\_TF 	&	RC\_S.	&	RC\_C. 	\\\hline
&&\multicolumn{19}{c|}{\textbf{Small Models} $(\leq 15B)$}\\\hline

Qwen 2	&	0.5	&	14.3	&	25.0	&	22.7	&	39.1	&	16.7	&	13.5	&	15.0	&	22.2	&	0.0	&	24.5	&	20.0	&	30.8	&	28.6	&	14.3	&	22.2	&	20.5	&	42.8	&	0.0	&	1.5	\\\hline
Llama 3.2	&	1	&	14.3	&	26.2	&	20.5	&	17.4	&	33.3	&	21.6	&	10.0	&	22.2	&	11.1	&	22.4	&	25.0	&	30.8	&	21.4	&	0.0	&	31.5	&	17.9	&	26.3	&	13.6	&	18.2	\\
Gemma 3	&	1	&	33.3	&	32.1	&	18.2	&	21.7	&	33.3	&	35.1	&	20.0	&	38.9	&	55.6	&	34.7	&	40.0	&	46.2	&	21.4	&	28.6	&	35.2	&	35.9	&	58.6	&	29.5	&	34.8	\\
Qwen 2	&	1.5	&	33.3	&	31.0	&	27.3	&	30.4	&	16.7	&	32.4	&	50.0	&	20.4	&	22.2	&	20.4	&	30.0	&	23.1	&	0.0	&	0.0	&	33.3	&	30.8	&	38.2	&	9.1	&	13.6	\\
StableLM 2	&	1.6	&	19.0	&	16.7	&	22.7	&	17.4	&	16.7	&	13.5	&	10.0	&	24.1	&	11.1	&	26.5	&	20.0	&	23.1	&	28.6	&	28.6	&	27.8	&	12.8	&	10.5	&	4.5	&	9.1	\\
EuroLLM	&	1.7	&	19.0	&	25.0	&	13.6	&	43.5	&	16.7	&	13.5	&	15.0	&	18.5	&	11.1	&	16.3	&	25.0	&	15.4	&	7.1	&	0.0	&	33.3	&	15.4	&	25.7	&	11.4	&	9.1	\\
Gemma 2	&	2	&	33.3	&	35.7	&	27.3	&	43.5	&	33.3	&	37.8	&	35.0	&	31.5	&	22.2	&	32.7	&	40.0	&	23.1	&	35.7	&	0.0	&	25.9	&	23.1	&	44.7	&	18.2	&	21.2	\\
Llama 3.2	&	3	&	28.6	&	34.5	&	27.3	&	34.8	&	33.3	&	18.9	&	15.0	&	27.8	&	44.4	&	24.5	&	20.0	&	38.5	&	21.4	&	14.3	&	35.2	&	20.5	&	39.5	&	22.7	&	30.3	\\
Phi 3	&	3.8	&	23.8	&	25.0	&	36.4	&	21.7	&	33.3	&	27.0	&	35.0	&	24.1	&	11.1	&	26.5	&	25.0	&	30.8	&	14.3	&	42.9	&	25.9	&	20.5	&	40.1	&	29.5	&	33.3	\\
Phi 3.5	&	3.8	&	9.5	&	17.9	&	11.4	&	26.1	&	0.0	&	16.2	&	20.0	&	24.1	&	0.0	&	22.4	&	20.0	&	30.8	&	14.3	&	42.9	&	18.5	&	12.8	&	20.4	&	20.5	&	16.7	\\
Gemma 3	&	4	&	33.3	&	51.2	&	50.0	&	30.4	&	50.0	&	35.1	&	20.0	&	29.6	&	22.2	&	28.6	&	25.0	&	46.2	&	7.1	&	14.3	&	33.3	&	30.8	&	44.1	&	61.4	&	40.9	\\
Qwen 2	&	7	&	33.3	&	38.1	&	34.1	&	30.4	&	50.0	&	37.8	&	40.0	&	25.9	&	33.3	&	28.6	&	25.0	&	30.8	&	21.4	&	42.9	&	25.9	&	41.0	&	45.4	&	40.9	&	37.9	\\
Llama 2	&	7	&	28.6	&	32.1	&	27.3	&	34.8	&	33.3	&	37.8	&	20.0	&	22.2	&	33.3	&	16.3	&	15.0	&	38.5	&	0.0	&	0.0	&	27.8	&	28.2	&	46.7	&	22.7	&	33.3	\\
WizardLM 2	&	7	&	19.0	&	33.3	&	40.9	&	34.8	&	16.7	&	40.5	&	30.0	&	25.9	&	22.2	&	26.5	&	25.0	&	15.4	&	50.0	&	42.9	&	29.6	&	41.0	&	35.5	&	15.9	&	34.8	\\
Mistral	&	7	&	33.3	&	38.1	&	25.0	&	21.7	&	16.7	&	27.0	&	35.0	&	27.8	&	22.2	&	24.5	&	35.0	&	15.4	&	14.3	&	28.6	&	24.1	&	30.8	&	42.8	&	25.0	&	37.9	\\
Aya-23	&	8	&	47.6	&	34.5	&	27.3	&	34.8	&	16.7	&	37.8	&	30.0	&	16.7	&	55.6	&	14.3	&	5.0	&	15.4	&	21.4	&	14.3	&	33.3	&	38.5	&	40.1	&	22.7	&	40.9	\\
Llama 3.1	&	8	&	33.3	&	29.8	&	34.1	&	21.7	&	16.7	&	37.8	&	40.0	&	25.9	&	22.2	&	28.6	&	30.0	&	38.5	&	7.1	&	28.6	&	35.2	&	30.8	&	45.4	&	45.5	&	42.4	\\
Llama 3	&	8	&	28.6	&	35.7	&	34.1	&	26.1	&	16.7	&	32.4	&	35.0	&	22.2	&	33.3	&	24.5	&	25.0	&	38.5	&	0.0	&	14.3	&	38.9	&	33.3	&	42.1	&	31.8	&	43.9	\\
DeepSeek-R1	&	8	&	33.3	&	44.0	&	31.8	&	52.2	&	0.0	&	27.0	&	35.0	&	40.7	&	11.1	&	40.8	&	35.0	&	38.5	&	42.9	&	28.6	&	38.9	&	20.5	&	42.1	&	43.2	&	45.5	\\
GLM 4	&	9	&	57.1	&	56.0	&	52.3	&	43.5	&	33.3	&	24.3	&	20.0	&	33.3	&	11.1	&	30.6	&	30.0	&	30.8	&	28.6	&	14.3	&	25.9	&	25.6	&	46.1	&	54.5	&	56.1	\\
EuroLLM	&	9	&	38.1	&	42.9	&	27.3	&	60.9	&	50.0	&	32.4	&	50.0	&	29.6	&	22.2	&	26.5	&	25.0	&	30.8	&	7.1	&	28.6	&	33.3	&	33.3	&	40.8	&	38.6	&	40.9	\\
Gemma 2	&	9	&	57.1	&	61.9	&	50.0	&	65.2	&	33.3	&	29.7	&	30.0	&	37.0	&	11.1	&	40.8	&	20.0	&	53.8	&	42.9	&	71.4	&	35.2	&	23.1	&	44.7	&	47.7	&	45.5	\\
Mistral-Nemo	&	12	&	28.6	&	42.9	&	31.8	&	17.4	&	16.7	&	24.3	&	30.0	&	29.6	&	22.2	&	28.6	&	25.0	&	23.1	&	35.7	&	42.9	&	29.6	&	15.4	&	40.8	&	27.3	&	36.4	\\
StableLM 2	&	12	&	23.8	&	28.6	&	25.0	&	30.4	&	16.7	&	8.1	&	5.0	&	3.7	&	11.1	&	4.1	&	5.0	&	0.0	&	0.0	&	14.3	&	25.9	&	7.7	&	30.9	&	27.3	&	24.2	\\
Gemma 3	&	12	&	47.6	&	65.5	&	56.8	&	73.9	&	33.3	&	29.7	&	25.0	&	25.9	&	33.3	&	24.5	&	30.0	&	23.1	&	21.4	&	28.6	&	42.6	&	20.5	&	53.9	&	77.3	&	60.6	\\
Llama 2	&	13	&	38.1	&	32.1	&	25.0	&	13.0	&	50.0	&	35.1	&	10.0	&	18.5	&	44.4	&	20.4	&	20.0	&	30.8	&	7.1	&	0.0	&	33.3	&	23.1	&	23.7	&	25.0	&	24.2	\\
Phi 3	&	14	&	19.0	&	35.7	&	34.1	&	30.4	&	16.7	&	35.1	&	50.0	&	29.6	&	33.3	&	26.5	&	20.0	&	30.8	&	14.3	&	42.9	&	40.7	&	33.3	&	40.8	&	29.5	&	39.4	\\
Phi 4	&	14	&	23.8	&	48.8	&	45.5	&	47.8	&	33.3	&	35.1	&	50.0	&	42.6	&	11.1	&	42.9	&	30.0	&	61.5	&	35.7	&	57.1	&	38.9	&	30.8	&	47.4	&	59.1	&	51.5	\\\hline
																																									
&&\multicolumn{19}{c|}{\textbf{Medium-sized Models} $(> 15B \& \leq 200B)$}\\\hline																																									
Mistral Small	&	22	&	28.6	&	50.0	&	36.4	&	26.1	&	16.7	&	16.2	&	25.0	&	35.2	&	33.3	&	38.8	&	35.0	&	46.2	&	42.9	&	42.9	&	27.8	&	15.4	&	48.0	&	38.6	&	50.0	\\
Gemma 2	&	27	&	47.6	&	66.7	&	52.3	&	60.9	&	50.0	&	35.1	&	45.0	&	29.6	&	11.1	&	28.6	&	25.0	&	30.8	&	21.4	&	42.9	&	46.3	&	35.9	&	51.3	&	77.3	&	72.7	\\
Gemma 3	&	27	&	38.1	&	75.0	&	56.8	&	87.0	&	50.0	&	54.1	&	55.0	&	50.0	&	44.4	&	51.0	&	55.0	&	38.5	&	35.7	&	57.1	&	61.1	&	46.2	&	61.8	&	70.5	&	69.7	\\
QWQ	&	32	&	47.6	&	66.7	&	45.5	&	60.9	&	33.3	&	40.5	&	40.0	&	42.6	&	11.1	&	42.9	&	45.0	&	38.5	&	42.9	&	71.4	&	35.2	&	35.9	&	40.8	&	70.5	&	63.6	\\
Aya-23	&	35	&	28.6	&	44.0	&	31.8	&	47.8	&	16.7	&	32.4	&	30.0	&	29.6	&	22.2	&	26.5	&	25.0	&	23.1	&	21.4	&	42.9	&	31.5	&	30.8	&	39.5	&	43.2	&	50.0	\\
Command-R	&	35	&	38.1	&	44.0	&	31.8	&	39.1	&	16.7	&	32.4	&	35.0	&	33.3	&	44.4	&	36.7	&	25.0	&	69.2	&	21.4	&	28.6	&	35.2	&	35.9	&	40.1	&	34.1	&	53.0	\\
Alfred	&	40	&	23.8	&	32.1	&	38.6	&	34.8	&	33.3	&	29.7	&	20.0	&	37.0	&	11.1	&	36.7	&	35.0	&	53.8	&	21.4	&	28.6	&	25.9	&	35.9	&	44.7	&	36.4	&	30.3	\\
Mixtral	&	8x7	&	28.6	&	46.4	&	50.0	&	43.5	&	16.7	&	27.0	&	30.0	&	24.1	&	22.2	&	26.5	&	25.0	&	38.5	&	14.3	&	42.9	&	35.2	&	23.1	&	37.5	&	34.1	&	51.5	\\
Llama 2	&	70	&	33.3	&	32.1	&	31.8	&	17.4	&	16.7	&	27.0	&	20.0	&	24.1	&	33.3	&	24.5	&	20.0	&	46.2	&	7.1	&	0.0	&	33.3	&	25.6	&	48.0	&	34.1	&	39.4	\\
Llama 3	&	70	&	52.4	&	58.3	&	45.5	&	56.5	&	33.3	&	35.1	&	25.0	&	31.5	&	33.3	&	30.6	&	25.0	&	38.5	&	14.3	&	28.6	&	37.0	&	33.3	&	42.8	&	50.0	&	48.5	\\
Llama 3.1	&	70	&	28.6	&	63.1	&	56.8	&	65.2	&	33.3	&	35.1	&	35.0	&	48.1	&	44.4	&	46.9	&	40.0	&	38.5	&	57.1	&	57.1	&	53.7	&	30.8	&	47.4	&	52.3	&	54.5	\\
DeepSeek-R1	&	70	&	47.6	&	73.8	&	56.8	&	47.8	&	16.7	&	29.7	&	40.0	&	46.3	&	11.1	&	49.0	&	55.0	&	38.5	&	42.9	&	42.9	&	51.9	&	17.9	&	46.1	&	65.9	&	66.7	\\
Qwen 2	&	72	&	33.3	&	56.0	&	50.0	&	47.8	&	0.0	&	40.5	&	35.0	&	38.9	&	33.3	&	38.8	&	30.0	&	38.5	&	57.1	&	57.1	&	35.2	&	33.3	&	45.4	&	54.5	&	60.6	\\
Command-R+	&	104	&	23.8	&	47.6	&	43.2	&	52.2	&	50.0	&	21.6	&	0.0	&	38.9	&	11.1	&	36.7	&	30.0	&	46.2	&	35.7	&	42.9	&	44.4	&	5.1	&	46.1	&	43.2	&	40.9	\\
Mistral-Large	&	123	&	42.9	&	75.0	&	56.8	&	56.5	&	50.0	&	45.9	&	50.0	&	40.7	&	33.3	&	40.8	&	35.0	&	61.5	&	21.4	&	14.3	&	37.0	&	35.9	&	44.1	&	68.2	&	50.0	\\
Mixtral	&	8x22	&	38.1	&	48.8	&	54.5	&	56.5	&	33.3	&	54.1	&	45.0	&	38.9	&	44.4	&	38.8	&	35.0	&	53.8	&	21.4	&	42.9	&	35.2	&	43.6	&	45.4	&	52.3	&	54.5	\\
WizardLM 2	&	8x22	&	28.6	&	47.6	&	36.4	&	26.1	&	16.7	&	29.7	&	35.0	&	37.0	&	33.3	&	34.7	&	40.0	&	53.8	&	14.3	&	0.0	&	48.1	&	23.1	&	46.1	&	52.3	&	51.5	\\\hline
																																									
&&\multicolumn{19}{c|}{\textbf{Large Models} $(> 200B)$}\\\hline																																									
																																									
Llama 3.1	&	405	&	42.9	&	67.9	&	68.2	&	47.8	&	33.3	&	37.8	&	40.0	&	40.7	&	22.2	&	36.7	&	30.0	&	53.8	&	28.6	&	14.3	&	55.6	&	33.3	&	55.9	&	63.6	&	59.1	\\
DeepSeek-R1	&	671	&	81.0	&	94.0	&	93.2	&	91.3	&	100.0	&	70.3	&	65.0	&	68.5	&	55.6	&	69.4	&	55.0	&	84.6	&	71.4	&	100.0	&	81.5	&	64.1	&	79.6	&	86.4	&	84.8	\\
ChatGPT 3.5	&	unk	&	71.4	&	91.7	&	90.9	&	100.0	&	50.0	&	73.0	&	60.0	&	75.9	&	44.4	&	77.6	&	75.0	&	92.3	&	71.4	&	71.4	&	75.9	&	69.2	&	85.5	&	88.6	&	83.3	\\
Claude 3.5 S	&	unk	&	42.9	&	85.7	&	88.6	&	100.0	&	50.0	&	70.3	&	75.0	&	90.7	&	66.7	&	91.8	&	85.0	&	100.0	&	92.9	&	85.7	&	85.2	&	69.2	&	88.2	&	86.4	&	86.4	\\
ChatGPT 4o	&	unk	&	52.4	&	86.9	&	90.9	&	95.7	&	66.7	&	78.4	&	60.0	&	75.9	&	55.6	&	77.6	&	70.0	&	92.3	&	78.6	&	85.7	&	74.1	&	64.1	&	83.6	&	86.4	&	83.3	\\
ChatGPT 4o m	&	unk	&	38.1	&	79.8	&	68.2	&	87.0	&	16.7	&	56.8	&	65.0	&	53.7	&	22.2	&	55.1	&	50.0	&	76.9	&	42.9	&	71.4	&	64.8	&	51.3	&	61.8	&	86.4	&	81.8	\\
Gemini 2.0 F	&	unk	&	33.3	&	81.0	&	84.1	&	87.0	&	16.7	&	70.3	&	70.0	&	75.9	&	66.7	&	79.6	&	75.0	&	84.6	&	71.4	&	85.7	&	68.5	&	53.8	&	80.9	&	86.4	&	84.8	\\
LeChat	&	unk	&	38.1	&	65.5	&	61.4	&	47.8	&	50.0	&	24.3	&	50.0	&	35.2	&	44.4	&	34.7	&	30.0	&	38.5	&	35.7	&	42.9	&	35.2	&	35.9	&	39.5	&	40.9	&	47.0	\\\hline

    \end{tabular}}}
    \caption{Complete results of experiments on fine-grained categories}
    \label{tab:app_detail_fine}
\end{table*}
\clearpage
\begin{table*}[h!]
    \centering
     \resizebox{\textwidth}{!}{
    \begin{tabular}{|l|l|r|}
    \hline
       Category name & Category Description& \#\\\hline					
V\_idiom	&	a question relating to idioms or metaphors 	&	21	\\
V\_noun	&	 a vocabulary question relating to a noun	&	84	\\
V\_verb	&	a vocabulary question relating to a verb	&	44	\\
V\_adjective	&	a vocabulary question relating to an adjective	&	23	\\
V\_native	&	a vocabulary question where the correct answer is a word exclusive to Luxembourgish 	&	6	\\
G\_gender	&	a question relating to determining the gender of a determinant, adjective, or pronoun 	&	37	\\
G\_pronouns	&	a question relating to pronouns	&	20	\\
G\_verbs	&	a question relating to conjugations of verbs	&	55	\\
G\_adjective	&	a question relating to adjectives	&	9	\\
G\_tense	&	a question relating to the correct use of tenses	&	50	\\
G\_past	&	a question relating to the past tense	&	21	\\
G\_present	&	a question relating to the present tense 	&	13	\\
G\_subjunctive	&	a question relating to the subjunctive tense	&	14	\\
G\_modal	&	a question relating to modal verbs 	&	7	\\
G\_preposition	&	a question relating to the correct use of prepositions or conjunctions 	&	54	\\
G\_cases	&	a question relating to nominative,genitive,dative, or accusative	&	39	\\
RC\_TF	&	a reading comprehension question where the answer is either true or false	&	152	\\
RC\_Simple	&	a reading comprehension question that does not require to understand the entire text excerpt to answer correctly	&	72	\\
RC\_Complex	&	a reading comprehension question requiring a deeper understanding of the entire text excerpt to answer correctly	&	86	\\\hline

    \end{tabular}}
    \caption{Statistics for fine-grained categories for exam questions.}
    \label{tab:app_cat_fine}
\end{table*}

\clearpage

\begin{table*}[]
    \centering
    \rotatebox{90}{
    \resizebox{1\textwidth}{!}{
    \begin{tabular}{|l|l||r||r|r|r|r||r|r|r|r||r|r|r|r||r|r|r|r|}
    \hline
	&		&		&	\multicolumn{8}{c||}{\textbf{Small Description}}				&														\multicolumn{8}{c|}{\textbf{News Headline}	}											\\\hline
\textbf{Model}	&	\textbf{Size}	&	\textbf{Exam Score}	&	\multicolumn{4}{c||}{\textbf{Grammar}}		&						\multicolumn{4}{c||}{\textbf{Adequacy}}		&						\multicolumn{4}{c||}{\textbf{Grammar}}			&					\multicolumn{4}{c|}{\textbf{Adequacy}}							\\\hline
			&		&	&Run 1	&	Run 2	&	Run 3	&	Average	&	Run 1	&	Run 2	&	Run 3	&	Average	&	Run 1	&	Run 2	&	Run 3	&	Average	&	Run 1	&	Run 2	&	Run 3	&	Average	\\\hline
		&	&		\multicolumn{17}{c|}{\textbf{Small Models} $(\leq 15B)$}\\\hline																																
Qwen 2	&	0.5	&	0.220	&	0.174	&	0.318	&	0.290	&	0.261	&	0.451	&	0.340	&	0.330	&	0.374	&	0.124	&	0.160	&	0.352	&	0.212	&	0.303	&	0.303	&	0.433	&	0.347	\\
Llama 3.2	&	1	&	0.044	&	0.534	&	0.520	&	0.582	&	0.545	&	0.521	&	0.572	&	0.550	&	0.548	&	0.338	&	0.256	&	0.418	&	0.337	&	0.228	&	0.257	&	0.525	&	0.337	\\
Gemma 3 & 1 & 0.005 &   0.700&	0.700&	0.800&	0.733&0.630&0.790&0.590&0.670 &0.300&0.250&0.010&0.187&0.350&0.290&0.260&0.300\\
Qwen 2	&	1.5	&	0.286	&	0.538	&	0.433	&	0.502	&	0.491	&	0.553	&	0.462	&	0.442	&	0.485	&	0.226	&	0.162	&	0.406	&	0.265	&	0.398	&	0.377	&	0.492	&	0.422	\\
StableLM 2	&	1.6	&	0.094	&	0.025	&	0.232	&	0.332	&	0.196	&	0.284	&	0.298	&	0.212	&	0.265	&	0.214	&	0.256	&	0.390	&	0.287	&	0.360	&	0.285	&	0.513	&	0.386	\\
EuroLLM & 1.7   &   0.006 &0.090&0.030&0.170&0.0967&0.220&0.370&0.130&0.240&0.100&0.220&0.000&0.107&0.210&0.210&0.220&0.213\\
Gemma 2	&	2	&	0.278	&	0.456	&	0.642	&	0.568	&	0.555	&	0.533	&	0.535	&	0.658	&	0.575	&	0.304	&	0.408	&	0.438	&	0.383	&	0.608	&	0.492	&	0.495	&	0.532	\\
Llama 3.2	&	3	&	0.321	&	0.830	&	0.782	&	0.812	&	0.808	&	0.689	&	0.737	&	0.648	&	0.691	&	0.692	&	0.702	&	0.656	&	0.683	&	0.653	&	0.647	&	0.695	&	0.665	\\
Phi 3	&	3.8	&	0.300	&	0.045	&	0.362	&	0.477	&	0.294	&	0.391	&	0.320	&	0.362	&	0.358	&	0.220	&	0.202	&	0.434	&	0.285	&	0.327	&	0.320	&	0.600	&	0.416	\\
Phi 3.5	&	3.8	&	0.101	&	0.267	&	0.258	&	0.287	&	0.271	&	0.243	&	0.150	&	0.147	&	0.180	&	0.118	&	0.065	&	0.488	&	0.224	&	0.240	&	0.328	&	0.625	&	0.398	\\
Gemma 3 & 4 &0.004 &0.760&0.720&0.810&0.763&0.650&0.800&0.630&0.693&0.650&0.260&0.020&0.310&0.550&0.320&0.270&0.380\\
Qwen 2	&	7	&	0.373	&	0.456	&	0.725	&	0.687	&	0.623	&	0.644	&	0.660	&	0.688	&	0.664	&	0.380	&	0.336	&	0.546	&	0.421	&	0.578	&	0.597	&	0.548	&	0.574	\\
Llama 2	&	7	&	0.141	&	0.054	&	0.275	&	0.238	&	0.189	&	0.296	&	0.275	&	0.225	&	0.265	&	0.146	&	0.180	&	0.424	&	0.250	&	0.425	&	0.408	&	0.522	&	0.452	\\
WizardLM 2	&	7	&	0.267	&	0.140	&	0.440	&	0.585	&	0.388	&	0.529	&	0.572	&	0.432	&	0.511	&	0.298	&	0.248	&	0.466	&	0.337	&	0.505	&	0.553	&	0.615	&	0.558	\\
Mistral	&	7	&	0.325	&	0.242	&	0.767	&	0.650	&	0.553	&	0.634	&	0.695	&	0.680	&	0.670	&	0.518	&	0.474	&	0.548	&	0.513	&	0.548	&	0.537	&	0.633	&	0.573	\\

Aya-23	&	8	&	0.322	&	0.377	&	0.470	&	0.708	&	0.518	&	0.510	&	0.515	&	0.448	&	0.491	&	0.086	&	0.142	&	0.446	&	0.225	&	0.340	&	0.372	&	0.557	&	0.423	\\
Llama 3.1	&	8	&	0.384	&	0.655	&	0.658	&	0.755	&	0.689	&	0.610	&	0.487	&	0.620	&	0.572	&	0.684	&	0.644	&	0.704	&	0.677	&	0.528	&	0.643	&	0.603	&	0.592	\\
Llama 3	&	8	&	0.357	&	0.624	&	0.690	&	0.722	&	0.678	&	0.630	&	0.643	&	0.627	&	0.633	&	0.322	&	0.310	&	0.554	&	0.395	&	0.600	&	0.578	&	0.630	&	0.603	\\
DeepSeek-R1	&	8	&	0.338	&	0.628	&	0.592	&	0.667	&	0.629	&	0.533	&	0.507	&	0.482	&	0.507	&	0.576	&	0.458	&	0.532	&	0.522	&	0.637	&	0.592	&	0.543	&	0.591	\\
GLM 4	&	9	&	0.421	&	0.721	&	0.783	&	0.795	&	0.766	&	0.591	&	0.590	&	0.715	&	0.632	&	0.322	&	0.252	&	0.402	&	0.325	&	0.580	&	0.658	&	0.578	&	0.606	\\
EuroLLM & 9 & 0.002 &0.250&0.560&0.300&	0.370&0.590&0.750&0.210&0.517&0.150&0.240&0.010&0.133&0.430&0.250&0.240&0.307\\
Gemma 2	&	9	&	0.360	&	0.622	&	0.807	&	0.818	&	0.749	&	0.623	&	0.645	&	0.762	&	0.677	&	0.738	&	0.584	&	0.652	&	0.658	&	0.730	&	0.678	&	0.550	&	0.653	\\
Mistral-Nemo	&	12	&	0.340	&	0.698	&	0.737	&	0.753	&	0.729	&	0.663	&	0.718	&	0.698	&	0.693	&	0.578	&	0.502	&	0.572	&	0.551	&	0.707	&	0.752	&	0.677	&	0.712	\\
StableLM 2	&	12	&	0.235	&	0.263	&	0.577	&	0.348	&	0.396	&	0.437	&	0.320	&	0.325	&	0.361	&	0.202	&	0.190	&	0.442	&	0.278	&	0.323	&	0.432	&	0.572	&	0.442	\\
Gemma 3 & 12 & 0.128 &0.800&0.850&0.820&0.823&0.670&0.820&0.740&0.743&0.780&0.270&0.130&0.393&0.680&0.330&0.290&0.433\\

Llama 2	&	13	&	0.147	&	0.140	&	0.438	&	0.315	&	0.298	&	0.286	&	0.230	&	0.295	&	0.270	&	0.200	&	0.236	&	0.512	&	0.316	&	0.460	&	0.468	&	0.583	&	0.504	\\
Phi 3	&	14	&	0.370	&	0.124	&	0.358	&	0.435	&	0.306	&	0.331	&	0.328	&	0.242	&	0.300	&	0.212	&	0.152	&	0.422	&	0.262	&	0.307	&	0.263	&	0.582	&	0.384	\\
Phi 4 & 14 & 0.028 & 0.170&	0.900&	0.800&	0.623&0.690&0.770&0.740&0.733&0.600&0.460&0.330&0.463&0.510&0.630&0.320&0.487\\
\hline
	&		&		\multicolumn{17}{c|}{\textbf{Medium-sized Models} $(> 15B \& \leq 200B)$}\\\hline																																
Mistral Small	&	22	&	0.024	&	0.675	&	0.843	&	0.848	&	0.789	&	0.633	&	0.620	&	0.692	&	0.648	&	0.812	&	0.772	&	0.738	&	0.774	&	0.750	&	0.780	&	0.620	&	0.717	\\
Gemma 2	&	27	&	0.066	&	0.648	&	0.853	&	0.882	&	0.794	&	0.663	&	0.728	&	0.808	&	0.733	&	0.776	&	0.674	&	0.730	&	0.727	&	0.780	&	0.760	&	0.607	&	0.716	\\
Gemma 3 & 27 & 0.245 &0.850&0.860&0.820&0.843&0.680&0.840&0.740&0.753&0.650&0.280&0.300&0.410&0.720&0.510&0.300&0.510\\
QWQ & 32 & 0.102 &0.190&0.150&0.800&0.380&0.700&0.630&0.490&0.607&0.500&0.470&0.530&0.500&0.250&0.300&0.370&0.307\\

Aya-23	&	35	&	0.036	&	0.711	&	0.620	&	0.667	&	0.666	&	0.631	&	0.697	&	0.573	&	0.634	&	0.450	&	0.374	&	0.508	&	0.444	&	0.558	&	0.558	&	0.625	&	0.581	\\
Command-R	&	35	&	0.021	&	0.517	&	0.653	&	0.745	&	0.638	&	0.621	&	0.658	&	0.667	&	0.649	&	0.182	&	0.423	&	0.642	&	0.416	&	0.617	&	0.688	&	0.670	&	0.658	\\
Alfred	&	40	&	0.011	&	0.206	&	0.402	&	0.503	&	0.370	&	0.381	&	0.420	&	0.405	&	0.402	&	0.272	&	0.350	&	0.272	&	0.298	&	0.495	&	0.573	&	0.342	&	0.470	\\
Mixtral	&	8x7	&	0.003	&	0.226	&	0.470	&	0.503	&	0.400	&	0.454	&	0.437	&	0.512	&	0.468	&	0.292	&	0.258	&	0.504	&	0.351	&	0.630	&	0.638	&	0.663	&	0.644	\\
Llama 2	&	70	&	0.005	&	0.464	&	0.632	&	0.592	&	0.562	&	0.599	&	0.505	&	0.623	&	0.576	&	0.152	&	0.212	&	0.482	&	0.282	&	0.432	&	0.495	&	0.622	&	0.516	\\
Llama 3	&	70	&	0.101	&	0.788	&	0.815	&	0.860	&	0.821	&	0.719	&	0.778	&	0.763	&	0.753	&	0.336	&	0.294	&	0.580	&	0.403	&	0.735	&	0.722	&	0.718	&	0.725	\\
Llama 3.1	&	70	&	0.121	&	0.842	&	0.865	&	0.895	&	0.867	&	0.731	&	0.818	&	0.800	&	0.783	&	0.712	&	0.730	&	0.810	&	0.751	&	0.777	&	0.800	&	0.700	&	0.759	\\
DeepSeek-R1	&	70	&	0.107	&	0.872	&	0.865	&	0.838	&	0.858	&	0.597	&	0.767	&	0.652	&	0.672	&	0.772	&	0.630	&	0.724	&	0.709	&	0.762	&	0.785	&	0.618	&	0.722	\\
Qwen 2	&	72	&	0.068	&	0.776	&	0.803	&	0.708	&	0.762	&	0.684	&	0.652	&	0.633	&	0.656	&	0.584	&	0.516	&	0.730	&	0.610	&	0.747	&	0.765	&	0.655	&	0.722	\\
Command-R+	&	104	&	0.057	&	0.705	&	0.688	&	0.818	&	0.737	&	0.627	&	0.683	&	0.700	&	0.670	&	0.646	&	0.682	&	0.754	&	0.694	&	0.697	&	0.777	&	0.693	&	0.722	\\
Mistral-Large	&	123	&	0.076	&	0.752	&	0.777	&	0.832	&	0.787	&	0.663	&	0.682	&	0.752	&	0.699	&	0.852	&	0.790	&	0.822	&	0.821	&	0.827	&	0.818	&	0.743	&	0.796	\\
Mixtral	&	8x22	&	0.011	&	0.515	&	0.622	&	0.790	&	0.642	&	0.566	&	0.567	&	0.528	&	0.554	&	0.620	&	0.604	&	0.700	&	0.641	&	0.780	&	0.782	&	0.700	&	0.754	\\
WizardLM 2	&	8x22	&	0.037	&	0.216	&	0.520	&	0.428	&	0.388	&	0.407	&	0.360	&	0.392	&	0.386	&	0.404	&	0.390	&	0.600	&	0.465	&	0.720	&	0.740	&	0.658	&	0.706	\\\hline
	&		&		\multicolumn{17}{c|}{\textbf{Large Models} $(> 200B)$}\\\hline																																
Llama 3.1	&	405	&	0.552	&	0.776	&	0.762	&	0.755	&	0.764	&	0.666	&	0.695	&	0.623	&	0.661	&	0.874	&	0.814	&	0.854	&	0.847	&	0.820	&	0.803	&	0.742	&	0.788	\\
DeepSeek-R1	&	671	&	0.804	&	0.980	&	0.857	&	0.890	&	0.909	&	0.630	&	0.705	&	0.760	&	0.698	&	0.866	&	0.658	&	0.653	&	0.726	&	0.800	&	0.817	&	0.647	&	0.754	\\
ChatGPT 3.5	&	unk	&	0.809	&	0.632	&	0.833	&	0.750	&	0.738	&	0.659	&	0.718	&	0.682	&	0.686	&	0.610	&	0.716	&	0.614	&	0.647	&	0.595	&	0.670	&	0.643	&	0.636	\\
Claude 3.5 Sonnet	&	unk	&	0.834	&	0.874	&	0.752	&	0.875	&	0.834	&	0.739	&	0.667	&	0.780	&	0.728	&	0.884	&	0.856	&	0.740	&	0.827	&	0.862	&	0.860	&	0.777	&	0.833	\\
ChatGPT 4o	&	unk	&	0.795	&	0.873	&	0.902	&	0.910	&	0.895	&	0.726	&	0.673	&	0.760	&	0.720	&	0.728	&	0.834	&	0.720	&	0.761	&	0.833	&	0.845	&	0.765	&	0.814	\\
ChatGPT 4o mini	&	unk	&	0.666	&	0.863	&	0.735	&	0.858	&	0.819	&	0.500	&	0.608	&	0.680	&	0.596	&	0.766	&	0.754	&	0.684	&	0.735	&	0.705	&	0.718	&	0.720	&	0.714	\\
Gemini 2.0 Flash	&	unk	&	0.768	&	0.814	&	0.792	&	0.815	&	0.807	&	0.606	&	0.555	&	0.512	&	0.557	&	0.544	&	0.550	&	0.512	&	0.535	&	0.698	&	0.777	&	0.675	&	0.717	\\
LeChat	&	unk	&	0.438	&	0.639	&	0.505	&	0.543	&	0.562	&	0.444	&	0.630	&	0.597	&	0.557	&	0.778	&	0.578	&	0.455	&	0.604	&	0.705	&	0.725	&	0.597	&	0.676	\\\hline

    \end{tabular}}}
    \caption{Detailed LuxGen Evaluation Results}
    \label{tab:app_luxgen}
\end{table*}\clearpage

\subsection{Comparison of LLM-as-a-Judge and traditional metrics}
\label{sec:app_comp}
\begin{figure*}
    \centering
    \includegraphics[width=\linewidth]{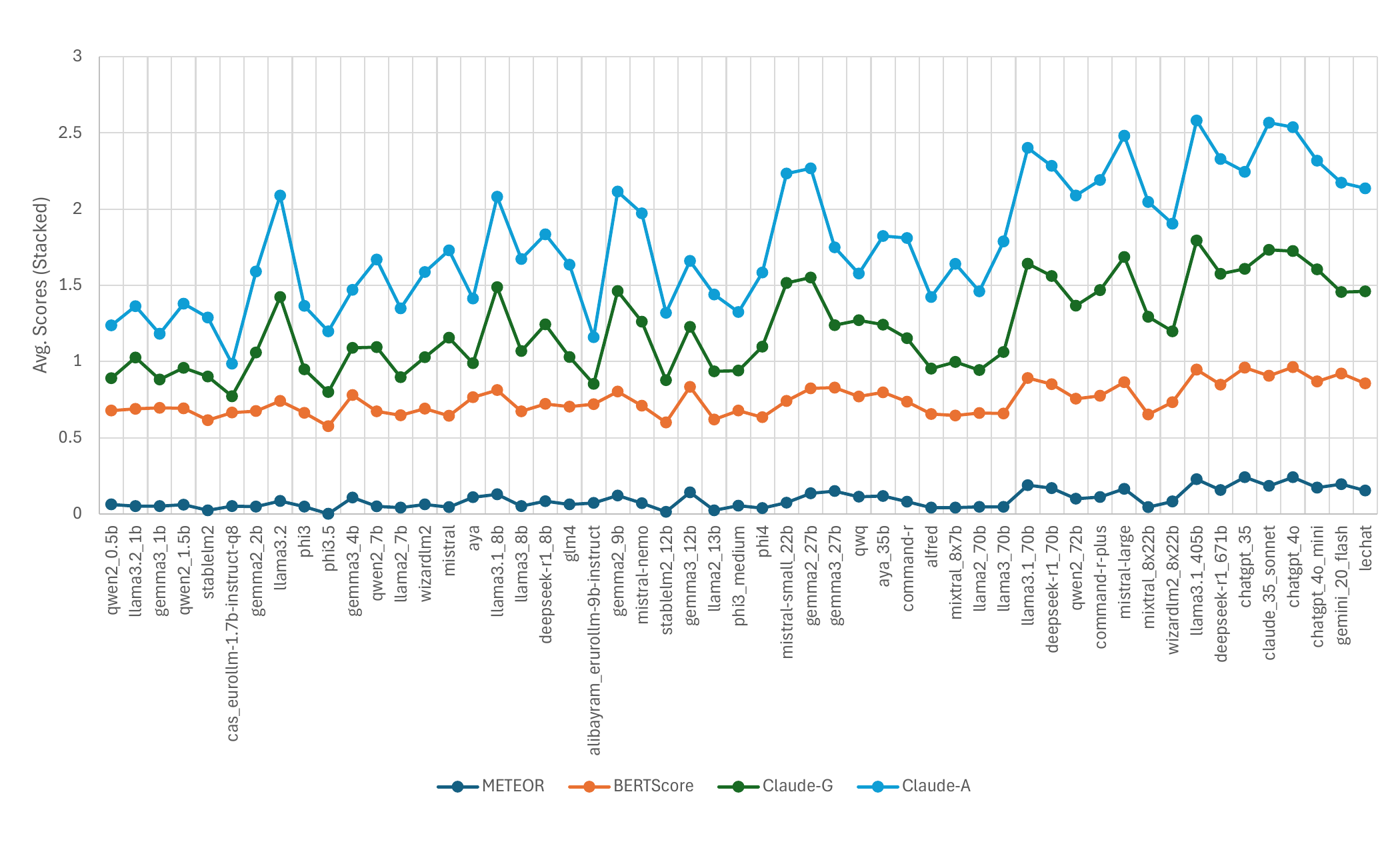}
    \caption{Comparison between LLM-as-a-Judge metrics, BERTScore, and METEOR on the headline generation task. We are using a stacked line chart to better highlight the correlations between the metrics.}
    \label{fig:comp_headlines}
\end{figure*}

\begin{figure*}
    \centering
    \includegraphics[width=\linewidth]{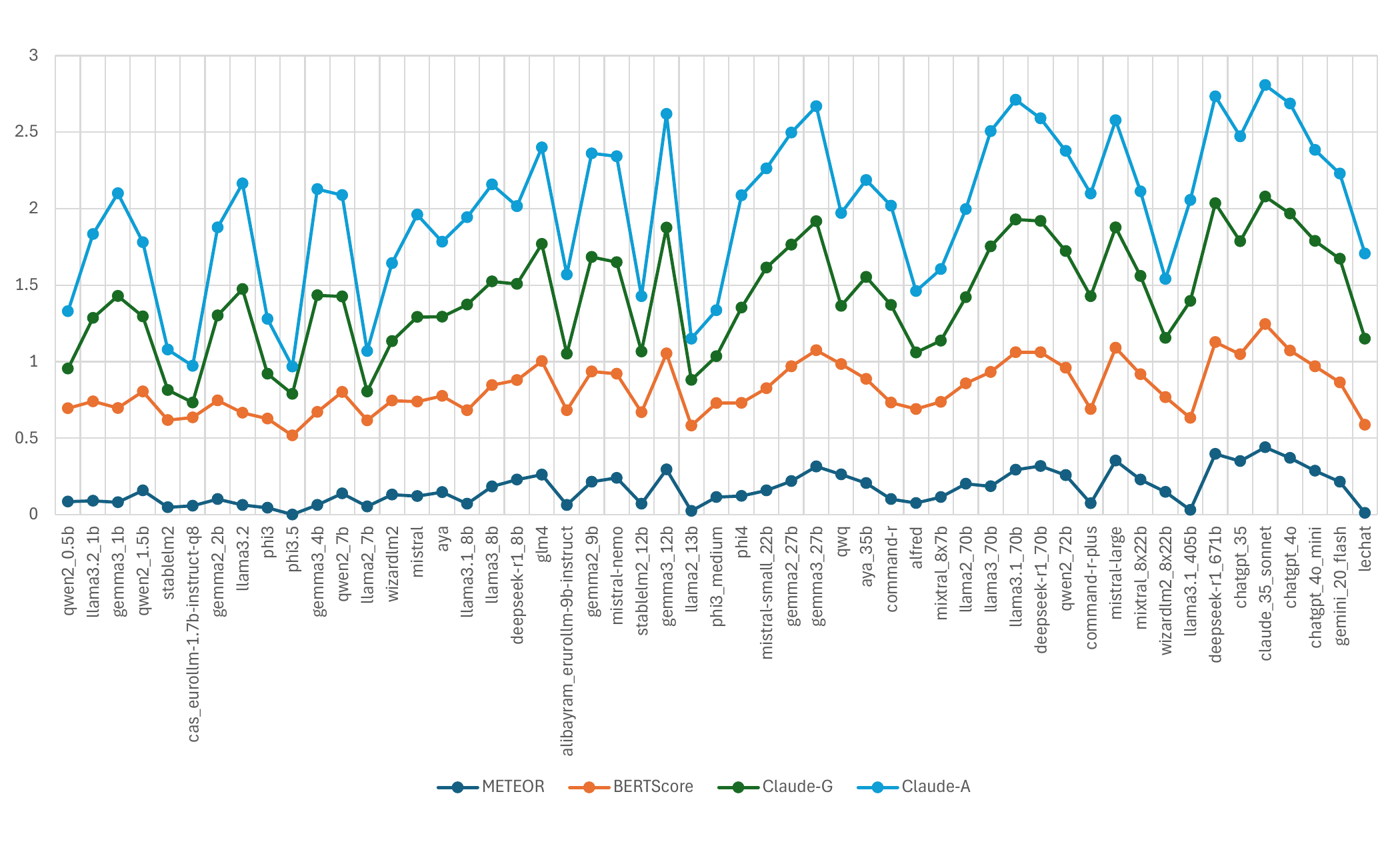}
    \caption{Comparison between LLM-as-a-Judge metrics, BERTScore, and METEOR on the short description task. We are using a stacked line chart to better highlight the correlations between the metrics.}
    \label{fig:comp_desc}
\end{figure*}

In this section, we compare our LLM-as-a-Judge metrics to the BERTScore and METEOR scores in order to assess the usefulness of our novel metrics. 
Figure~\ref{fig:comp_headlines} and \ref{fig:comp_desc} show performances of every studied LLM on the Headline Generation and Short Description tasks using the four different metrics, respectively. We observe that while the scores are different for each metric they appear to be highly correlated to each other. 
This is further corroborated by Tables~\ref{tab:comp_headlines1} through \ref{tab:comp_desc3} which contain correlation matrices with Pearson Correlation Coefficients, t-scores, and p-values between every metric, showing strong correlations between the traditional metrics and the LLM-as-a-Judge metrics, with the significance tests showing the correlations to be significant at a significance level $\alpha = 0.01$.

    \begin{table}[]
        \centering
        \begin{subtable}[b]{\linewidth}
             \resizebox{\textwidth}{!}{
   
        \begin{tabular}{|c|c|c|c|c|}
        \hline
	&	Claude-G	&	Claude-A	&	METEOR	&	BERTScore	\\\hline
Claude-G	&	1.00	&	0.86	&	0.71	&	0.70	\\
Claude-A	&	0.86	&	1.00	&	0.56	&	0.55	\\
METEOR	&	0.71	&	0.56	&	1.00	&	0.95	\\
BERTScore	&	0.70	&	0.55	&	0.95	&	1.00	\\\hline

        \end{tabular}}
        \caption{Pearson Correlation Coefficient}
        \label{tab:comp_headlines1}
                \end{subtable}

    \begin{subtable}[b]{\linewidth}
        \centering
            \resizebox{\textwidth}{!}{

        \begin{tabular}{|c|c|c|c|c|}
        \hline
	&	Claude-G	&	Claude-A	&	METEOR	&	BERTScore	\\\hline
Claude-G	&	/	&	11.90	&	7.27	&	7.07	\\
Claude-A	&	11.90	&	/	&	4.82	&	4.72	\\
METEOR	&	7.27	&	4.82	&	/	&	22.60	\\
BERTScore	&	7.07	&	4.72	&	22.60	&	/	\\\hline

        \end{tabular}}
        \caption{T-Scores}
        \end{subtable}
        \label{tab:comp_headlines2}

    \begin{subtable}[b]{\linewidth}
        \centering
     \resizebox{\textwidth}{!}{

        \begin{tabular}{|c|c|c|c|c|}
        \hline
	&	Claude-G	&	Claude-A	&	METEOR	&	BERTScore	\\\hline
Claude-G	&	/	&	2.472E-16	&	2.050E-09	&	4.241E-09	\\
Claude-A	&	2.472E-16	&	/	&	1.322E-05	&	1.859E-05	\\
METEOR	&	2.051E-09	&	1.322E-05	&	/	&	3.134E-28	\\
BERTScore	&	4.241E-09	&	1.859E-05	&	3.134E-28	&	/	\\\hline

        \end{tabular}}
                \label{tab:comp_headlines3}

\caption{P-Values}

        \end{subtable} 
        \caption{Correlation tests for different metrics on the headline generation task.}
        \label{tab:comp_headlines}

    \end{table}
\begin{table}
\centering
    \begin{subtable}[b]{\linewidth}
        \centering
        \resizebox{\textwidth}{!}{
        \begin{tabular}{|c|c|c|c|c|}
        \hline
	&	Claude-G	&	Claude-A	&	METEOR	&	BERTScore	\\\hline
Claude-G	&	1.00	&	0.90	&	0.65	&	0.72	\\
Claude-A	&	0.90	&	1.00	&	0.60	&	0.73	\\
METEOR	&	0.65	&	0.60	&	1.00	&	0.89	\\
BERTScore	&	0.72	&	0.73	&	0.89	&	1.00	\\\hline

        \end{tabular}}
        \caption{Pearson Correlation Coefficient}
        \label{tab:comp_desc1}
    \end{subtable}

    \begin{subtable}[b]{\linewidth}
        \centering
        \resizebox{\textwidth}{!}{
        \begin{tabular}{|c|c|c|c|c|}
        \hline
	&	Claude-G	&	Claude-A	&	METEOR	&	BERTScore	\\\hline
Claude-G	&	/	&	15.08	&	6.13	&	7.36	\\
Claude-A	&	15.08	&	/	&	5.33	&	7.67	\\
METEOR	&	6.13	&	5.33	&	/	&	14.31	\\
BERTScore	&	7.36	&	7.67	&	14.31	&	/	\\\hline

        \end{tabular}}
        \caption{T-Scores}
        \label{tab:comp_desc2}
    \end{subtable}

    \begin{subtable}[b]{\linewidth}
        \centering
        \resizebox{\textwidth}{!}{
        \begin{tabular}{|c|c|c|c|c|}
        \hline
	&	Claude-G	&	Claude-A	&	METEOR	&	BERTScore	\\\hline
Claude-G	&	/	&	1.941E-20	&	1.266E-07	&	1.466E-09	\\
Claude-A	&	1.941E-20	&	/	&	2.215E-06	&	4.740E-10	\\
METEOR	&	1.266E-07	&	2.215E-06	&	/	&	1.736E-19	\\
BERTScore	&	1.466E-09	&	4.740E-10	&	1.736E-19	&	/	\\\hline

        \end{tabular}}
        \caption{P-Values}
        \label{tab:comp_desc3}

        \end{subtable} 
        \caption{Correlation tests for different metrics on the Short Description task.}
    \end{table}
\clearpage
\subsection{Results for the Short Description Task}
\label{sec:app_desc}

\begin{figure*}[h]
\centering
\begin{subfigure}[]{.49\textwidth}
\centering
    	    \includegraphics[width=\textwidth]{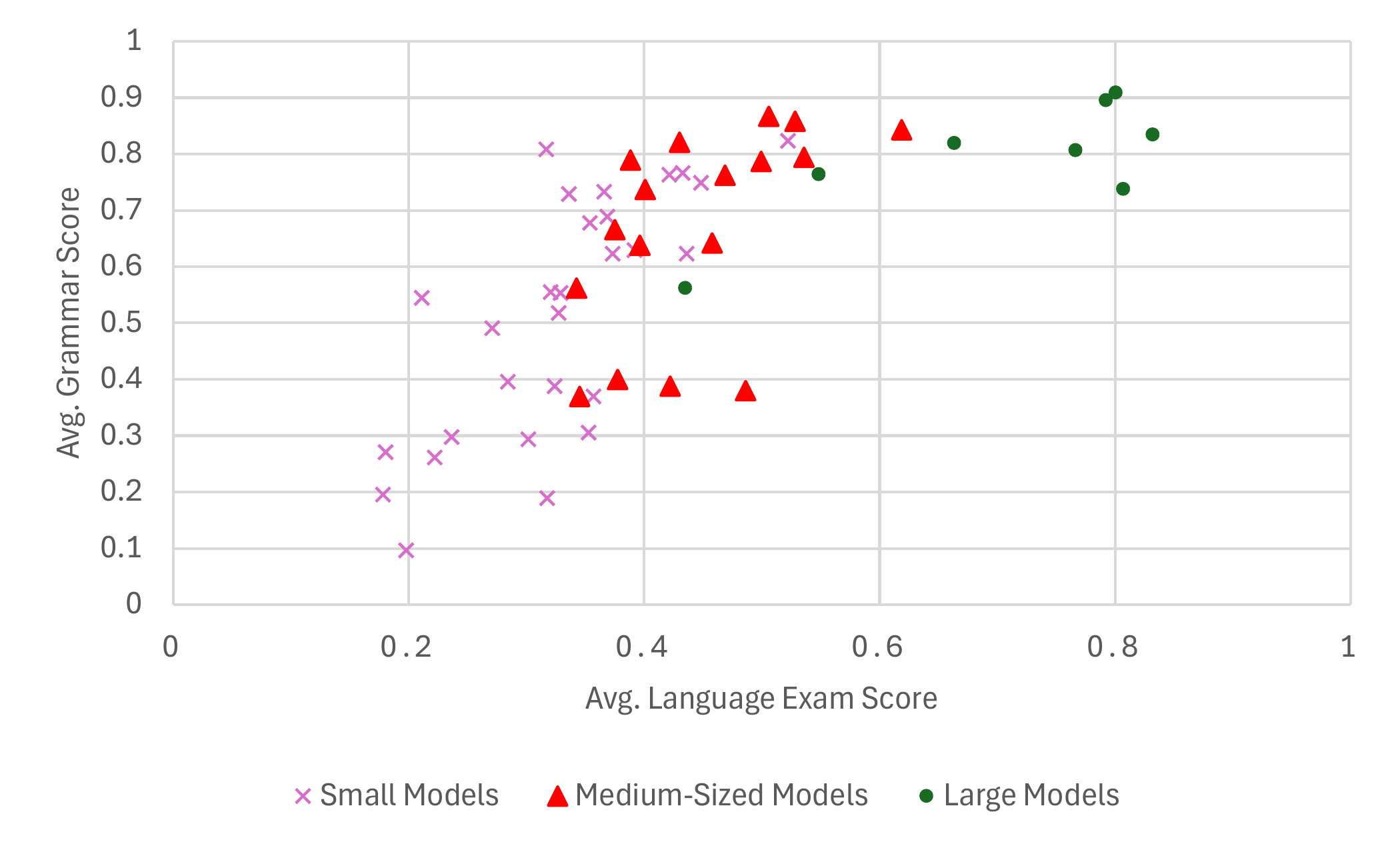}
            \caption{Claude Score - Grammar and Spelling}
            \label{fig:desc_gram}
\end{subfigure}
~
\begin{subfigure}[]{.49\textwidth}
\centering
            \includegraphics[width=\textwidth]{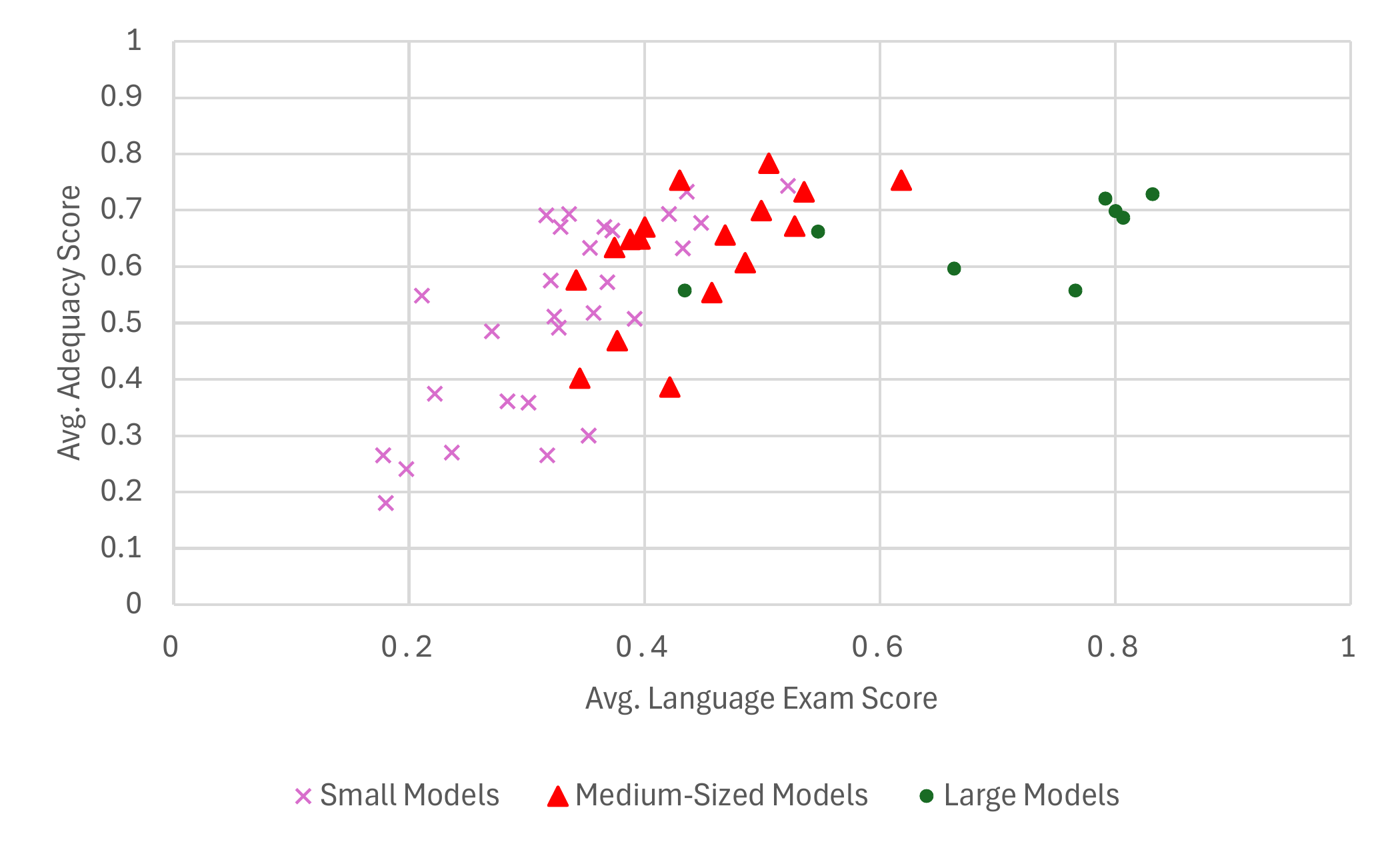}
            \caption{Claude Score - Adequacy}
            \label{fig:desc_ade}
\end{subfigure}

\centering
\begin{subfigure}[]{.49\textwidth}
\centering
    	    \includegraphics[width=\textwidth]{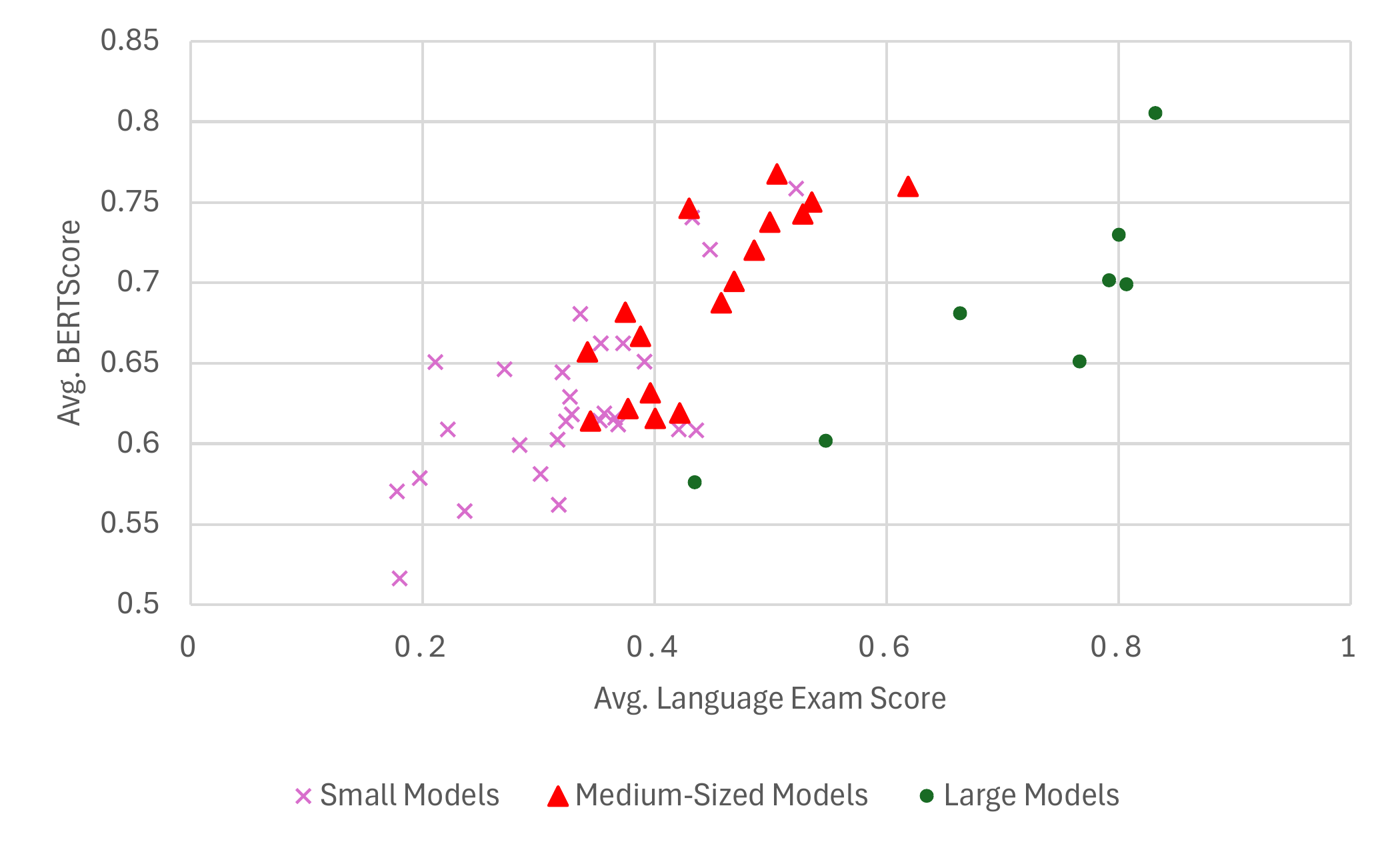}
            \caption{BERTScore}
            \label{fig:desc_bert}
\end{subfigure}
~
\vspace{-0.2cm}
\begin{subfigure}[]{.49\textwidth}
\centering
            \includegraphics[width=\textwidth]{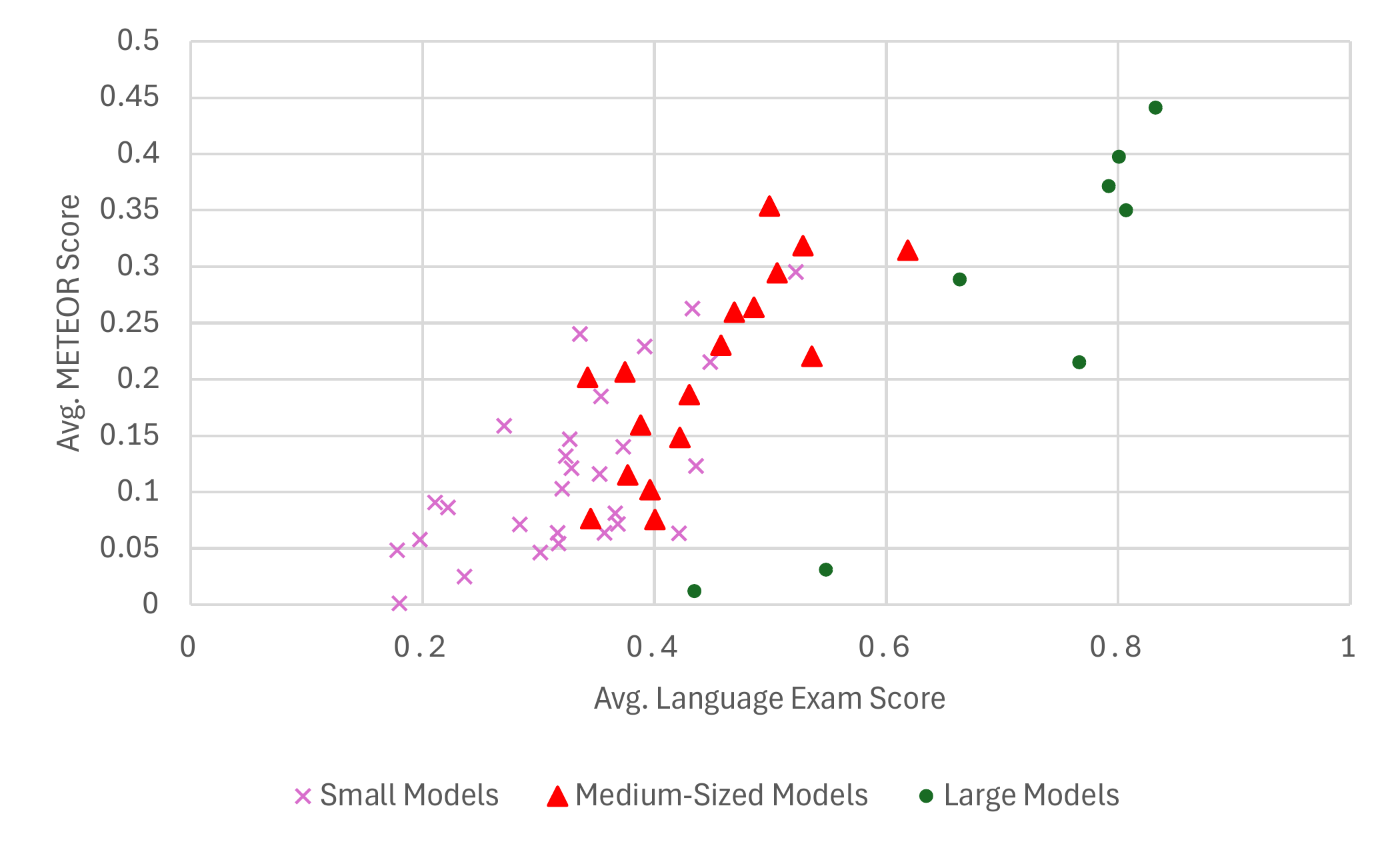}
            \caption{METEOR Score}
            \label{fig:desc_meteor}
\end{subfigure}
\caption{Performance on Language Exams vs Performance on the Short Description Task.}
\vspace{-.35cm}
\label{fig:luxgen_desc}
\end{figure*}

Figure~\ref{fig:luxgen_desc} shows the LLM performances on the Short Description task plotted against their performances on the proficiency exams while Table~\ref{tab:pearson_desc} shows PCCs, t-scores, and corresponding p-values. 

\begin{table}[h]
    \centering
    \begin{subtable}[b]{\linewidth}
    
    \resizebox{\textwidth}{!}{
    \begin{tabular}{|l|r|r|r|r|}
    \hline
Task	&	Small	&	Medium	&	Large	&	All	\\\hline
Claude-G	&	0.670	&	0.773	&	0.926	&	0.779	\\
Claude-A	&	0.700	&	0.812	&	0.859	&	0.660	\\
METEOR	&	0.743	&	0.550	&	0.780	&	0.702	\\
BERTScore	&	0.749	&	0.595	&	0.618	&	0.604	\\\hline

    \end{tabular}}
    \caption{Pearson Correlation Coefficients}
    \end{subtable}
    \begin{subtable}[b]{\linewidth}
    
    \resizebox{\textwidth}{!}{
    \begin{tabular}{|l|r|r|r|r|}
    \hline
		&	small	&	medium	&	large	&	all	\\\hline
Claude-G	&	6.443	&	8.702	&	17.516	&	8.884	\\
Claude-A	&	7.005	&	9.935	&	11.970	&	6.279	\\
METEOR	&	7.935	&	4.700	&	8.911	&	7.036	\\
BERTScore	&	8.072	&	5.293	&	5.612	&	5.419	\\\hline

    \end{tabular}}
    \caption{T-Scores}
    \end{subtable}
   \begin{subtable}[b]{\linewidth}
    
    \resizebox{\textwidth}{!}{
    \begin{tabular}{|l|r|r|r|r|}
    \hline
	&	small	&	medium	&	large	&	all	\\\hline
Claude-G	&	4.126E-08	&	1.182E-11	&	3.185E-23	&	6.204E-12	\\
Claude-A	&	5.338E-09	&	1.609E-13	&	1.980E-16	&	7.483E-08	\\
METEOR	&	1.837E-10	&	2.009E-05	&	5.635E-12	&	4.770E-09	\\
BERTScore	&	1.119E-10	&	2.563E-06	&	8.25E-07	&	1.64E-06	\\\hline

    \end{tabular}}
    \caption{P-Values}
    \end{subtable}
   
    \caption{Correlation tests between the overall exam scores and Short Description task performances.}
    \vspace{-.34cm}
    \label{tab:pearson_desc}
\end{table}

